# A Model-Based Active Testing Approach to Sequential Diagnosis

**Alexander Feldman**  A.B.FELDMAN@TUDELFT.NL
*Delft University of Technology*
*Mekelweg 4, 2628 CD, Delft, The Netherlands*

**Gregory Provan**  G.PROVAN@CS.UCC.IE
*University College Cork*
*Department of Computer Science*
*College Road, Cork, Ireland*

**Arjan van Gemund**  A.J.C.VANGEMUND@TUDELFT.NL
*Delft University of Technology*
*Mekelweg 4, 2628 CD, Delft, The Netherlands*

## Abstract

Model-based diagnostic reasoning often leads to a large number of diagnostic hypotheses. The set of diagnoses can be reduced by taking into account extra observations (passive monitoring), measuring additional variables (probing) or executing additional tests (sequential diagnosis/test sequencing). In this paper we combine the above approaches with techniques from Automated Test Pattern Generation (ATPG) and Model-Based Diagnosis (MBD) into a framework called Fractal (FRamework for ACtive Testing ALgorithms). Apart from the inputs and outputs that connect a system to its environment, in active testing we consider additional input variables to which a sequence of test vectors can be supplied. We address the computationally hard problem of computing optimal control assignments (as defined in Fractal) in terms of a greedy approximation algorithm called Fractal$^G$. We compare the decrease in the number of remaining minimal cardinality diagnoses of Fractal$^G$ to that of two more Fractal algorithms: Fractal$^{ATPG}$ and Fractal$^P$. Fractal$^{ATPG}$ is based on ATPG and sequential diagnosis while Fractal$^P$ is based on probing and, although not an active testing algorithm, provides a baseline for comparing the lower bound on the number of reachable diagnoses for the Fractal algorithms. We empirically evaluate the trade-offs of the three Fractal algorithms by performing extensive experimentation on the ISCAS85/74XXX benchmark of combinational circuits.

## 1. Introduction

Combinational Model-Based Diagnosis (MBD) approaches (de Kleer & Williams, 1987) often lead to a large number of diagnoses, which is exponential in the number of components, in the worst-case. Combining multiple sensor readings (observation vectors) (Pietersma & van Gemund, 2006) helps in a limited number of cases because the approach is inherently passive, i.e., there are situations in which the observations repeat themselves (for example, in systems that are stationary, pending a reconfiguration).

Sequential diagnosis algorithms (Shakeri, 1996) can be used as an alternative to the above passive approach, with better decay of the number of diagnostic hypotheses. The decay rate depends on the tests and test dictionary matrix, and is bounded from below by results for tests with binary outcomes. Algorithms for sequential diagnosis suffer from a number of other limitations. Early approaches assume single-faults while multiple-fault sequential diagnosis is super-exponential ($\Sigma_2^p$ or harder) (Shakeri, Raghavan, Pattipati, & Patterson-Hine, 2000).

As observations (test outcomes) are not known in advance, the goal of a diagnostician is to create a policy that minimizes the diagnostic uncertainty on average, i.e., one aims at minimizing the average depth of a test tree. Pattipati and Alexandridis (1990) have shown that under certain conditions (e.g., unit test costs and equal prior fault probabilities) a one-step look-ahead policy leads to an optimal average depth of the test tree; de Kleer, Raiman, and Shirley (1992) have shown that one-step look-ahead delivers good practical results for a range of combinational circuits.

This paper proposes a framework, called FRACTAL (FRamework for ACtive Testing ALgorithms) for comparing different computational vs. optimality trade-offs in various techniques for reducing the diagnostic uncertainty. All FRACTAL algorithms start from an initial set of multiple-fault diagnostic hypotheses (this initial set can contain all possible hypotheses) and compute actions for reducing this initial set to, if possible, a single diagnostic hypothesis (candidate). In the case of probing (de Kleer & Williams, 1987), this action consists of measuring an internal (hidden) variable. In the case of sequential diagnosis (ATPG), the action consists of applying a set of input (control) assignments that disambiguate the health state of the component that appears faulty in most of the initial diagnostic hypotheses. In the case of active testing the action consists of applying a set of input (control) assignments that optimally reduce the initial set of hypotheses. In our framework the active testing and sequential diagnosis approaches differ only in how they compute the input (control) assignments. We measure the optimality of the algorithms by computing the speed with which they decay the initial set of hypotheses and the computational efficiency.

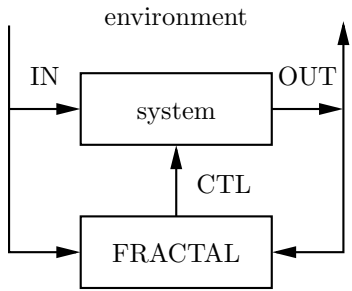

Figure 1: Active testing dataflow for FRACTAL

In FRACTAL, we study the influence of not just the input (IN) and output (OUT) variables, but also control (CTL) variables. Controls are similar to inputs, except they can be modified by users while the system is connected to its environment. Use of models from first principles and controls in FRACTAL allows us to eliminate the need of designing explicit tests and test dictionaries. Our algorithms implicitly create test matrices leading to optimal decay based on the built-in testing capabilities of the system. We call the above approach active testing; it is a technique using models from first principles and controls for creating test sequences that reduce the diagnostic uncertainty of a system. The architecture in which we use FRACTAL and active testing is shown in Fig. 1.

As reliable component failure rates may be problematic to obtain, we assume equally likely and small prior probabilities of failure and measure the diagnostic uncertainty as the number of Minimal Cardinality (MC) diagnoses. FRACTAL can be modified to use arbitrary failure probabilities and even components that are more likely to be faulty than healthy. This would necessitate modifications of some of the algorithms (e.g., change of bias in the importance sampling, etc.). In addition to simplifying the modeling, the equiprobable failure rates assumption also has computational advantages. It can be shown that with equal and small prior probabilities of failure, the diagnostic entropy, e.g., as used by de Kleer and Williams (1987), can be computed directly from the number of MC diagnoses.

The computational complexity of deterministic algorithms for sequential diagnosis increases with respect to both the fault-cardinality and the number of tests (the size of the test dictionary). To enable performance to scale up to real-world problems, which may have high fault-cardinality and a large number of tests, we propose FRACTAL$^G$–a low-cost greedy stochastic approach that maintains exponential decay of the number of MC diagnoses. Instead of assuming single faults or timing out, FRACTAL$^G$ may result in suboptimal but still exponential decay.

We study the performance of FRACTAL$^G$ compared to two alternatives: (1) FRACTAL$^{ATPG}$, which implements sequential diagnosis based on Automated Test Pattern Generation (ATPG), and (2) FRACTAL$^P$, which implements probing (de Kleer & Williams, 1987). ATPG has been successfully used in the electronic industry to compute sets of inputs that test each component in a VLSI circuit. We have considered an ATPG-based approach because it is natural to attempt to reduce the diagnostic ambiguity by computing inputs that can disambiguate the status of the single component that appears in the majority of the diagnostic hypotheses.

FRACTAL$^{ATPG}$ is derived from sequential testing, is deterministic and myopic, and allows us to evaluate how well a single-step lookahead approach works on the given model. Although probing is not classified as a technique for sequential diagnosis, it can be viewed as a process for generating tests using additional control circuitry (machine or human) to execute a probe such that some output reveals the internal variable. Its significance is that it shows a lower bound on the number of diagnoses achievable for a model extended with unlimited CTL circuitry.

Our contributions are as follows:

- We devise an approach for reducing the diagnostic uncertainty, called active testing, that generalizes sequential diagnosis and MBD, allows combination of multiple passive sensor readings, and does not require explicit tests and test dictionaries.

- We design FRACTAL$^{ATPG}$—a single-step look-ahead algorithm based on ATPG—for solving the active testing problem.

- We design and implement FRACTAL$^G$–a greedy approximation algorithm for active testing that overcomes the limitations of FRACTAL$^{ATPG}$ and offers a trade-off in computational complexity vs. optimality for reducing the diagnostic uncertainty. We compare FRACTAL$^G$ and FRACTAL$^{ATPG}$.

- We implement FRACTAL$^P$ and use it as a computationally efficient, myopic (one-step lookahead), easy-to-analyze baseline technique for reducing diagnostic uncertainty.

Although FRACTAL$^P$ is technically not an active testing algorithm, the implementation of probing and active testing in a common framework and the unified experimentation help to understand the cost vs. performance trade-offs in (active and passive) testing vs. probing strategies.

- We present extensive empirical data on 74XXX/ISCAS85 circuits, which enable us to evaluate FRACTAL$^{ATPG}$, FRACTAL$^G$, and FRACTAL$^P$ in terms of their ability to reduce the number of remaining diagnoses according to a geometric decay function.

This paper is organized as follows. Section 2 introduces related work. Section 3 presents basic MBD notions, the concept of remaining number of diagnoses and a framework for sequential diagnosis. Section 4 introduces a stochastic sampling-based algorithm for computing the expected number of cardinality-minimal diagnoses. Section 5 describes the FRACTAL$^{ATPG}$, FRACTAL$^G$, and FRACTAL$^P$ algorithms. Section 6 shows experimental results. Finally, Sec. 7 summarizes this paper and discusses future work.

## 2. Related Work

Early work aimed at diagnostic convergence by de Kleer and Williams (1987) compute a probe sequence for reducing diagnostic entropy using a myopic search strategy. Unlike their work, in active testing we assume that probes are not available, other than indirectly exposed through diagnosis based on test vectors, which offers an automated solution.

Generating test vectors to deduce faults has received considerable attention. Automatic test pattern generation (ATPG) aims at verifying particular, single-faults (Stephan, Brayton, & Sangiovanni-Vincentelli, 1996). ATPG differs from active testing in that the vectors are specific for particular single-faults, whereas active testing generates a sequence of vectors to isolate *unknown, multiple*-faults, a much harder problem.

Table 1: Properties of techniques for sequential diagnosis

| Technique | Model | User Actions | Automatic Tests | Performance | Cost |
|---|---|---|---|---|---|
| Passive monitoring | first principles | - | - | variable[1] | low |
| Sequential diagnosis | test dictionary | apply test | no | good | variable[2] |
| FRACTAL$^{ATPG}$ | first principles | apply controls | yes | variable[3] | medium |
| FRACTAL$^G$ | first principles | apply controls | yes | good | high |
| Probing (FRACTAL$^P$) | first principles | measure internals | - | binary search | medium |

[1] Depends on the environment (IN/OUT data).
[2] Speed deteriorates rapidly with multiple-faults.
[3] Depends on the model topology.

Table 1 summarizes the properties of the various techniques for sequential diagnosis discussed in this paper. FRACTAL eliminates the need for using tools for building tests and test dictionaries, such as the ones proposed by Deb, Ghoshal, Malepati, and Kleinman (2000). In our approach tests and test dictionaries are automatically constructed from design speci-fications and models. At the same time, FRACTAL delivers comparable or better diagnostic convergence at reasonable computational price.

Active testing bears some resemblance with sequential diagnosis, which also generates a sequence of test vectors (Pattipati & Alexandridis, 1990; Raghavan, Shakeri, & Pattipati, 1999; Tu & Pattipati, 2003; Kundakcioglu & Ünlüyurt, 2007). The principal difference is that in sequential diagnosis a fault dictionary is used ("fault matrix"). This pre-compiled dictionary has the following drawback: in order to limit the (exponential) size of the dictionary, the number of stored test vectors is extremely small compared to the test vector space. This severely constrains the optimality of the vector sequence that can be generated; in contrast, active testing computes arbitrary test vectors on the fly using a model-based approach. Furthermore, the matrix specifies tests that only have a binary (pass/fail) outcome, whereas active testing exploits all the system's outputs, leading to faster diagnostic convergence. In addition, we allow the inputs to be dynamic, which makes our framework suitable for online fault isolation.

The sequential diagnosis problem studies optimal trees when there is a cost associated with each test (Tu & Pattipati, 2003). When costs are equal, it can be shown that the optimization problem reduces to a next best control problem (assuming one uses information entropy). In this paper a diagnostician who is given a sequence $S$ and who tries to compute the *next* optimal control assignment would try to minimize the expected number of remaining diagnoses $|\Omega(S)|$.

Our task is harder than that of Raghavan et al. (1999), since the diagnosis task is NP-hard, even though the diagnosis lookup uses a fault dictionary; in our case we compute a new diagnosis after every test. Hence we have an NP-hard sequential problem interleaved with the complexity of diagnostic inference at each step (in our case the complexity of diagnosis is $\Sigma_2^p$-hard). Apart from the above-mentioned differences, we note that optimal test sequencing is infeasible for the size of problems in which we are interested.

Model-Based Testing (MBT) (Struss, 1994) is a generalization of sequential diagnosis. The purpose of MBT is to compute inputs manifesting a certain (faulty) behavior. The main differences from our active testing approach are that MBT (1) assumes that all inputs are controllable and (2) MBT aims at *confirming* single-fault behavior as opposed to maximally decreasing the diagnostic uncertainty.

Brodie, Rish, Ma, and Odintsova (2003) cast their models in terms of Bayesian networks. Our notion of entropy is the size of the diagnosis space, whereas Brodie et al. use decision-theoretic notions of entropy to guide test selection. Brodie et al. extend their past Bayesian diagnostic approach (Rish, Brodie, & Ma, 2002) with sequential construction of probe sets (probe sets are collections of, for example, pings to a subset of the nodes in a computer network). The approach of Brodie et al. is limited to networks although it can be extended by modifying the type of Bayesian network shown by Rish et al.; such a modification, however, would necessitate more computationally expensive Bayesian reasoning for achieving good approximation results for the most probable explanations.

The approach of Brodie et al. (2003) does not compute modifications in the target network topology and does not propose control actions (for example, a network server that fails to respond can be dialed-up through a modem or checked by a technician at a higher cost). The similarity between FRACTAL and active probing is that both approaches attempt at reducing the diagnostic uncertainty by analyzing the future state of the system as a function of some action (sending a set of probes for active probing or an application of control inputs for Fractal$^{\text{G}}$ and Fractal$^{\text{ATPG}}$).

We solve a different problem than that of Heinz and Sachenbacher (2008), Alur, Courcoubetis, and Yannakakis (1995). Both of these approaches assume a non-deterministic model defined as an automaton. In contrast, our framework assumes a static system (plant model) for which we must compute a temporal sequence of tests to best isolate the diagnosis.

Esser and Struss (2007) also adopt an automaton framework for test generation, except that, unlike Heinz and Sachenbacher (2008) or Alur et al. (1995), they transform this automaton to a relational specification, and apply their framework to software diagnosis. This automaton-based framework accommodates more general situations than does ours, such as the possibility that the system's state after a transition may not be uniquely determined by the state before the transition and the input, and/or the system's state may be associated with several possible observations. In our MBD framework, a test consists of an instantiation of several variables, which corresponds to the notion of test sequence within the automaton framework of Heinz and Sachenbacher. The framework of Esser and Struss requires modeling of the possible faults, whereas Fractal works both with weak and strong-fault models[1]. Interestingly, as shown by Esser and Struss, modeling of abnormal software behavior can be derived to some extent from software functional requirements. This makes their framework suitable for software systems.

A recent approach to active diagnosis is described by Kuhn, Price, de Kleer, Do, and Zhou (2008), where additional test vectors are computed to optimize the diagnosis while the system (a copier) remains operational. Their work differs from ours in that plans (roughly analogous to test sequences) with a probability of failure $T$ are computed statically, and a plan remains unmodified even if it fails to achieve its desired goal (a manifestation of a failure with probability close to $T$). Conversely, Fractal dynamically computes next-best control settings in a game-like manner. The biggest difference between Fractal and the approach of Kuhn et al. is in the use of models. Fractal is compatible with traditional MBD (de Kleer & Williams, 1987) and can reuse existing models from first principles while the pervasive approach of Kuhn et al. uses an automaton and a set of possible actions.

The approach of Kuhn et al. (2008) uses existing MBD and planning algorithms, and as such integrates existing approaches; in contrast, Fractal introduces new control algorithms and reuses an external diagnostic oracle. An advantage of the pervasive diagnosis approach is that the use of a planning engine generates a complete sequence of actions, as opposed to the one-step lookahead of Fractal$^{\text{G}}$. Depending on the planning formalism, the complexity of pervasive diagnosis can be dominated by the planning module, while the most complex computational task in Fractal is that of diagnosis. Both pervasive diagnosis and this paper, however, report good average-case computational efficiency for benchmark problems. Last, the paper of Kuhn et al. is limited to single-fault diagnoses, although the pervasive diagnosis framework can be generalized to multiple faults.

Feldman, Provan, and van Gemund (2009a) introduce an early version of Fractal$^{\text{G}}$. This paper (1) generalizes the Fractal framework, (2) introduces Fractal$^{\text{ATPG}}$ and Fractal$^{\text{P}}$, (3) extends the experimental results, and (4) provides a comparison of the different Fractal approaches.

---

1. Weak-fault models (also known as models with ignorance of abnormal behavior) and strong-fault models are discussed by Feldman, Provan, and van Gemund (2009b).

## 3. Concepts and Definitions

Our discussion starts by introducing relevant MBD notions. Central to MBD, a *model* of an artifact is represented as a propositional **Wff** over a set of variables $V$. We will define four subsets of these variables: *assumable*, *observable*[2], *control*, and *internal* variables. This gives us our initial definition:

**Definition 1** (Active Testing System). An active testing system ATS is defined as ATS = $\langle$SD, COMPS, CTL, OBS$\rangle$, where SD is a propositional **Wff** over a variable set $V$, COMPS$\cup$ OBS $\cup$ CTL $\subseteq V$, and COMPS, OBS, and CTL are subsets of $V$ containing assumable, observable, and control variables, respectively.

The set of internal variables is denoted as INT, INT $= V \setminus \{$COMPS $\cup$ OBS $\cup$ CTL$\}$. Throughout this paper we assume that OBS, COMPS, and CTL are disjoint, and SD $\not\models \bot$. Sometimes it is convenient (but not necessary) to split OBS into non-controllable inputs IN and outputs OUT (OBS $=$ IN $\cup$ OUT, IN $\cap$ OUT $= \emptyset$).

### 3.1 Running Example

We will use the Boolean circuit shown in Fig. 2 as a running example for illustrating all notions and the algorithm shown in this paper. The 2-to-4 line demultiplexer consists of four Boolean inverters and four and-gates.

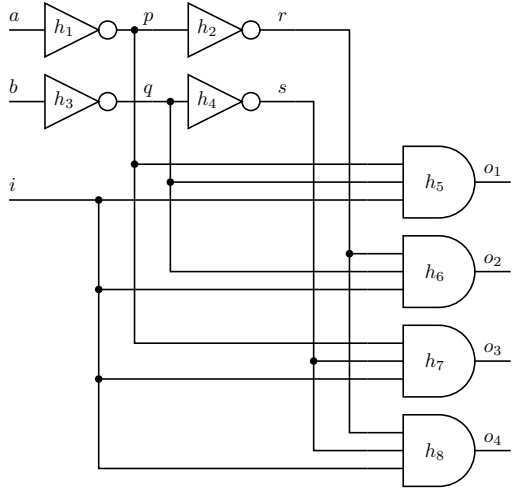

Figure 2: A demultiplexer circuit

The expression $h \Rightarrow (o \Leftrightarrow \neg i)$ models an inverter, where the variables $i$, $o$, and $h$ represent input, output, and health respectively. Similarly, an and-gate is modeled as $h \Rightarrow (o \Leftrightarrow i_1 \wedge i_2 \wedge i_3)$. The above propositional formulae are copied for each gate in Fig. 2 and their variables subscripted and renamed in such a way as to ensure a proper disambiguation

---

2. In the MBD literature the assumable variables are also referred to as "component", "failure-mode", or "health" variables. Observable variables are also called "measurable" variables.

and to connect the circuit. The result is the following propositional model:

$$\text{SD} = \begin{cases} [h_1 \Rightarrow (a \Leftrightarrow \neg p)] \wedge [h_2 \Rightarrow (p \Leftrightarrow \neg r)] \\ [h_3 \Rightarrow (b \Leftrightarrow \neg q)] \wedge [h_4 \Rightarrow (q \Leftrightarrow \neg s)] \\ h_5 \Rightarrow (o_1 \Leftrightarrow i \wedge p \wedge q) \\ h_6 \Rightarrow (o_2 \Leftrightarrow i \wedge r \wedge q) \\ h_7 \Rightarrow (o_3 \Leftrightarrow i \wedge p \wedge s) \\ h_8 \Rightarrow (o_4 \Leftrightarrow i \wedge r \wedge s) \end{cases} \quad (1)$$

The set of assumable variables is COMPS = $\{h_1, h_2, \ldots, h_8\}$, the observable variables are OBS = $\{a, b, o_1, o_2, o_3, o_4\}$, and the set of control variables is the singleton CTL = $\{i\}$. Note the conventional selection of the sign of the "health" variables $h_1, h_2, \ldots, h_n$. Other authors use "ab" for abnormal.

### 3.2 Diagnosis

The traditional query in MBD computes terms of assumable variables, which are explanations for the system description and an observation.

**Definition 2** (Diagnosis). Given a system ATS, an observation $\alpha$ over some variables in OBS, and an assignment $\omega$ to all variables in COMPS, $\omega$ is a diagnosis iff SD $\wedge \alpha \wedge \omega \not\models \bot$.

The set of all diagnoses of SD and an observation $\alpha$ is denoted as $\Omega(\text{SD}, \alpha)$. The cardinality of a diagnosis, denoted as $|\omega|$, is defined as the number of negative literals in $\omega$.

Continuing our running example, consider an observation vector $\alpha_1 = \neg a \wedge \neg b \wedge i \wedge o_4$. There are a total of 256 possible assignments to all variables in COMPS and $|\Omega(\text{SD}, \alpha_1)| = 200$. Example diagnoses are $\omega_1 = h_1 \wedge h_2 \wedge \ldots \wedge h_7 \wedge \neg h_8$ and $\omega_2 = \neg h_1 \wedge h_2 \wedge h_3 \wedge \neg h_4 \wedge h_5 \wedge h_6 \wedge h_7 \wedge h_8$. We will write sometimes a diagnosis in a set notation, specifying the set of negative literals only. Thus $\omega_2$ would be represented as $D_2 = \{\neg h_1, \neg h_4\}$.

**Definition 3** (Minimal-Cardinality Diagnosis). A diagnosis $\omega^\leq$ is defined as Minimal-Cardinality (MC) if no diagnosis $\tilde{\omega}^\leq$ exists such that $|\tilde{\omega}^\leq| < |\omega^\leq|$.

Our selection of minimality criterion is such that it is impossible to compute all diagnoses from the set of all MC diagnoses without further inference. MC diagnoses, however, are often used in practice due to the prohibitive cost of computing a representation of all diagnoses of a system and an observation (e.g., all subset-minimal diagnoses).

Consider an observation vector $\alpha_2 = \neg a \wedge \neg b \wedge i \wedge \neg o_1 \wedge o_4$. There are 6 MC diagnoses of cardinality 2 consistent with SD $\wedge \alpha_2$, and counting these MC diagnoses is a common problem in MBD.

The number of MC diagnoses of a system ATS and an observation $\alpha$ is denoted as $|\Omega^\leq(\text{SD}, \alpha)|$, where $\Omega^\leq(\text{SD}, \alpha)$ is the set of all MC diagnoses of SD $\wedge \alpha$. Given a system ATS, an observation sequence $S$ is defined as a $k$-tuple of terms $S = \langle \alpha_1, \alpha_2, \ldots, \alpha_k \rangle$, where $\alpha_i$ ($1 \leq i \leq k$) is an instantiation of variables in OBS.

Throughout this paper, we assume that the health of the system under test does not change during the test (i.e., the same inputs and a fault produce the same outputs) and call this assumption stationary health.

**Lemma 1.** *Given a system* ATS, *a stationary health state for its components* $\omega$, *and an observation sequence* $S$, *it follows that* $\omega \in \Omega(\mathrm{SD}, \alpha_1) \cap \Omega(\mathrm{SD}, \alpha_2) \cap \ldots \cap \Omega(\mathrm{SD}, \alpha_k)$.

*Proof.* The above statement follows immediately from the stationary health assumption and Def. 2. □

Lemma 1 can be applied only in the cases in which *all* diagnoses are considered. If we compute subset-minimal diagnoses in a weak-fault model, for example, the intersection operator has to be redefined to handle subsumptions. To handle non-characterizing sets of diagnoses[3] (e.g., MC or first $m$ diagnoses), we provide the following definition.

**Definition 4** (Consistency-Based Intersection). Given a set of diagnoses $D$ of $\mathrm{SD} \wedge \alpha$, and an a posteriori observation $\alpha'$, the intersection of $D$ with the diagnoses of $\mathrm{SD} \wedge \alpha'$, denoted as $\Omega^{\cap}(D, \alpha')$, is defined as the set $D'$ ($D' \subseteq D$) such that for each $\omega \in D'$ it holds that $\mathrm{SD} \wedge \alpha' \wedge \omega \not\models \bot$.

It is straightforward to generalize the above definition to an observation sequence $S$.

**Definition 5** (Remaining Minimal-Cardinality Diagnoses). Given a diagnostic system ATS and an observation sequence $S$, the set of remaining diagnoses $\Omega(S)$ is defined as $\Omega(S) = \Omega^{\cap}(\Omega^{\cap}(\cdots \Omega^{\cap}(\Omega^{\leq}(\mathrm{SD}, \alpha_1), \alpha_2), \cdots), \alpha_k)$.

We use $|\Omega(S)|$ instead of the more precise diagnostic entropy as defined by de Kleer and Williams (1987) and subsequent works, as this allows low-complexity estimations (discussed in Sec. 4). In particular, if all diagnoses are of minimal-cardinality and the failure probability of each component is the same, then the gain in the diagnostic entropy can be directly computed from $|\Omega(S)|$.

## 4. Computing the Expected Number of MC Diagnoses

Active testing aims to minimize the expected number of diagnoses that result from the possible set of outputs that may occur from a given control vector. In this section we present an algorithm to approximate this expectation.

We will compute the expected number of diagnoses for a set of observable variables $M$ ($M \subseteq \mathrm{OBS}$). The initial observation $\alpha$ and the set of MC diagnoses $D = \Omega^{\leq}(\mathrm{SD}, \alpha)$ modify the probability density function of subsequent outputs (observations), i.e., a subsequent observation $\alpha'$ changes its likelihood. The (non-normalized) a posteriori probability of an observation $\alpha'$, given a function $\Omega^{\leq}$ that computes the set of MC diagnoses and an initial observation $\alpha$, is:

$$\Pr(\alpha'|\mathrm{SD}, \alpha) = \frac{|\Omega^{\cap}(\Omega^{\leq}(\mathrm{SD}, \alpha), \alpha')|}{|\Omega^{\leq}(\mathrm{SD}, \alpha)|} \qquad (2)$$

The above formula computes the probability of a given a priori set of diagnoses restricting the possible outputs, i.e., we assume that the probability is the ratio of the number of remaining diagnoses to the number of initial diagnoses. In practice, there are many $\alpha$ for which $\Pr(\alpha'|\text{SD},\alpha) = 0$, because a certain fault heavily restricts the possible outputs of a system (i.e., the set of the remaining diagnoses in the numerator is empty).

---

3. A characterizing set of diagnoses, for example the set of all subset-minimal diagnoses, is loosely defined as a set of diagnoses from which the (complete) set of all diagnoses can be constructed without using the system description or any other information.

The expected number of remaining MC diagnoses for a variable set $M$, given an initial observation $\alpha$, is then the weighted average of the intersection sizes of all possible instantiations over the variables in $M$ (the weight is the probability of an output):

$$E^{\leq}(\text{SD},M|\alpha) = \frac{\displaystyle\sum_{\alpha' \in M^*} |\Omega^{\cap}(D,\alpha')| \cdot \Pr(\alpha'|\text{SD},\alpha)}{\displaystyle\sum_{\alpha' \in M^*} \Pr(\alpha'|\text{SD},\alpha)} \tag{3}$$

where $D = \Omega^{\leq}(\text{SD},\alpha)$ and $M^*$ is the set of all possible assignments to the variables in $M$. Replacing (2) in (3) and simplifying gives us the following definition:

**Definition 6** (Expected Minimal-Cardinality Diagnoses Intersection Size). Given a system ATS and an initial observation $\alpha$, the expected remaining number of MC diagnoses $E^{\leq}(\text{SD},\text{OBS}|\alpha)$ is defined as:

$$E^{\leq}(\text{SD},\text{OBS}|\alpha) = \frac{\displaystyle\sum_{\alpha' \in \text{OBS}^*} |\Omega^{\cap}(\Omega^{\leq}(\text{SD},\alpha),\alpha')|^2}{\displaystyle\sum_{\alpha' \in \text{OBS}^*} |\Omega^{\cap}(\Omega^{\leq}(\text{SD},\alpha),\alpha')|} \tag{4}$$

where $\text{OBS}^*$ is the set of all possible assignments to all variables in OBS.

Two of the algorithms presented in this paper compute the expected number of remaining MC diagnoses for one variable. As a result the expectation expression in (4) simplifies to:

$$E^{\leq}(\text{SD},v|\alpha) = \frac{|\Omega^{\cap}(\Omega^{\leq}(\text{SD},\alpha),v)|^2 + |\Omega^{\cap}(\Omega^{\leq}(\text{SD},\alpha),\neg v)|^2}{|\Omega^{\cap}(\Omega^{\leq}(\text{SD},\alpha),v)| + |\Omega^{\cap}(\Omega^{\leq}(\text{SD},\alpha),\neg v)|} \tag{5}$$

The complexity of computing (5) depends only on the length of the sequence $S$, the complexity of the MC oracle computing $\Omega^{\leq}(\text{SD},\alpha)$, and the complexity of the intersection algorithm.

### 4.1 Computing the Expectation Using Importance Sampling

To overcome the computational complexity of evaluating an expectation, we employ a stochastic algorithm based on importance sampling. The key insight that allows us to build a fast method for computing the expected number of remaining diagnoses is that the prior observation (and respectively the set of MC diagnoses) shifts the probability of the outputs. Hence, an algorithm that samples the possible input assignments (recall that it is a basic modeling assumption that inputs are equally likely) and counts the number of *different* observations, given the set of prior diagnoses, can produce a good approximation.

We next introduce an algorithm for approximating the expected number of remaining diagnoses.

---

**Algorithm 1** Approximate expectation of $\Omega(S)$
---
1: **function** EXPECTATION(ATS, $\gamma$, $D$) **returns** a real
    **inputs:** ATS (active testing system): model
              $\gamma$ (term): control vector
              $D$ (set of diagnoses): prior diagnoses
    **local variables:** $\alpha, \beta, \omega$ (terms): observation
                    $s$ (integer): sum of the remaining diagnoses, initially 0
                    $q$ (integer): sum of squares of the remaining diagnoses, initially 0
                    $Z$ (set of terms): samples
                    $\hat{E}$ (real): expectation
2:     $Z \leftarrow \emptyset$
3:     **repeat**
4:         $\alpha \leftarrow$ RANDOMINPUTS(SD, IN)
5:         **for all** $\omega \in D$ **do**
6:             $\beta \leftarrow$ INFEROUTPUTS(SD, OUT, $\alpha \wedge \gamma, \omega$)
7:             **if** $\alpha \wedge \beta \notin Z$ **then**
8:                 $Z \leftarrow Z \cup \{\alpha \wedge \beta\}$
9:                 $q \leftarrow q + |\Omega^\cap(D, \alpha \wedge \beta \wedge \gamma)|^2$
10:               $s \leftarrow s + |\Omega^\cap(D, \alpha \wedge \beta \wedge \gamma)|$
11:               $\hat{E} \leftarrow q/s$
12:             **end if**
13:         **end for**
14:     **until** TERMINATE($\hat{E}$)
15:     **return** $\hat{E}$
16: **end function**

---

Algorithm 1 uses a couple of auxiliary functions: RANDOMINPUTS assigns random values to all inputs and INFEROUTPUTS computes all outputs from the system model, all inputs and a diagnosis.[4] The computation of the intersection size $|\Omega^\cap(D, \alpha \wedge \beta \wedge \gamma)|$ can be implemented by counting those $\omega \in D$ for which SD $\wedge\, \alpha \wedge \beta \wedge \gamma \wedge \omega \not\models \bot$.

The algorithm terminates when a termination criterion (checked by TERMINATE) is satisfied. In our implementation, TERMINATE returns success when the last $n$ iterations (where $n$ is a small constant) leave the expected number of diagnoses, $\hat{E}$, unchanged, in terms of its integer representation. Our experiments show that for all problems considered, $n < 100$ yields a negligible error.

The complexity of Alg. 1 is determined by the complexity of consistency checking (line 9 – 10) and the size of $D$. If we denote the complexity of a single consistency check with $\Upsilon$, then the complexity of Alg. 1 becomes $O(|D|\Upsilon)$. Although consistency checking for diagnostic problems is *NP*-hard in the worst case, for average-case problems it is easy. In our implementation of EXPECTATION we overcome the complexity of consistency checking

---

4. This is not always possible in the general case. In our framework, we have a number of assumptions, i.e., a weak-fault model, well-formed circuit, etc. The complexity of INFEROUTPUTS thus depends on the framework and the assumptions.

by using an incomplete Logic-Based Truth Maintenance System (LTMS) (Forbus & de Kleer, 1993).

## 5. Algorithms for Reducing the Diagnostic Uncertainty

In this section we introduce three algorithms: FRACTAL$^{\text{ATPG}}$, FRACTAL$^{\text{G}}$, and FRACTAL$^{\text{P}}$.

### 5.1 Problem Definition and Exhaustive Search

Our AT problem is defined as follows:

**Problem 1** (Optimal Control Sequence). Given a system ATS, a sequence (of past observations and controls) $S = \langle \alpha_1 \wedge \gamma_1, \alpha_2 \wedge \gamma_2, \cdots, \alpha_k \wedge \gamma_k \rangle$, where $\alpha_i$ ($1 \leq i \leq k$) are OBS assignments and $\gamma_j$ ($1 \leq j \leq k$) are CTL assignments, compute a new CTL assignment $\gamma_{k+1}$, such that:

$$\gamma_{k+1} = \operatorname*{argmin}_{\gamma \in \text{CTL}^\star} E^{\leq}(\Omega^{\cap}(\text{SD}, S), \{\text{IN} \cup \text{OUT}\} | \gamma) \tag{6}$$

where $\text{CTL}^\star$ is the space of all possible control assignments.

Problem 1 is different from the general sequential testing problem, as formulated by Shakeri (1996). In the Shakeri formulation, there are different test costs and different prior failure probabilities, where Problem 1 assumes equal costs and equal small prior probabilities of failure. Pattipati and Alexandridis (1990) show that under those assumptions, minimizing the test cost at each step constitutes an optimal policy for minimizing the expected test cost. Hence, solving Problem 1 is solving the lesser problem of generating an optimal test strategy given unit costs and equal prior failure probability. Note that we can use an algorithm that optimizes Problem 1 as a heuristic algorithm for solving the sequential testing problem. In this case the expected cost would be arbitrarily far from the optimum one, depending on the cost distribution and the tests.

Consider our running example with an initial observation vector (and control assignment) $\alpha_3 \wedge \gamma_3 = a \wedge b \wedge i \wedge o_1 \wedge \neg o_2 \wedge \neg o_3 \wedge \neg o_4$, where $\gamma_3 = i$ is chosen as the initial control input. The four MC diagnoses of $\text{SD} \wedge \alpha_3 \wedge \gamma_3$ are $\Omega^{\leq} = \{\{\neg h_1, \neg h_3\}, \{\neg h_2, \neg h_5\}, \{\neg h_4, \neg h_5\}, \{\neg h_5, \neg h_8\}\}$.

An exhaustive algorithm would compute the expected number of diagnoses for each of the $2^{|\text{CTL}|}$ next possible control assignments. In our running example we have one control variable $i$ and two possible control assignments ($\gamma_5 = i$ and $\gamma_6 = \neg i$). To compute the expected number of diagnoses, for each possible control assignment $\gamma$ and for each possible observation vector $\alpha$, we have to count the number of initial diagnoses which are consistent with $\alpha \wedge \gamma$.

Computing the intersection sizes for our running example gives us Table 2. Note that, in order to save space, Table 2 contains rows only for those $\alpha \wedge \gamma$ for which $\Pr(\alpha \wedge \gamma) \neq 0$, given the initial diagnoses $\Omega^{\leq}$ (and, as a result, $|\Omega^{\cap}(\Omega^{\leq}(\text{SD}, \alpha_3 \wedge \gamma_3), \alpha \wedge \gamma)| \neq 0$). It is straightforward to compute the expected number of diagnoses for any control assignment with the help of this marginalization table. In order to do this we have to (1) filter out those lines which are consistent with the control assignment $\gamma$ and (2) compute the sum and the sum of the squares of the intersection sizes (the rightmost column of Table 2).

Table 2: Marginalization table for SD and $\alpha_3$

| $i$ | $a$ | $b$ | $o_1$ | $o_2$ | $o_3$ | $o_4$ | Pr | $|\Omega^\cap|$ | $i$ | $a$ | $b$ | $o_1$ | $o_2$ | $o_3$ | $o_4$ | Pr | $|\Omega^\cap|$ |
|---|---|---|---|---|---|---|---|---|---|---|---|---|---|---|---|---|---|
| F | F | F | F | F | F | F | 0.03125 | 1 | F | T | T | T | F | F | T | 0.03125 | 1 |
| F | F | F | T | F | F | F | 0.0625 | 2 | T | F | F | F | F | F | T | 0.0625 | 2 |
| F | F | F | T | F | F | T | 0.03125 | 1 | T | F | F | F | F | T | F | 0.03125 | 1 |
| F | F | T | F | F | F | F | 0.03125 | 1 | T | F | F | F | T | F | F | 0.03125 | 1 |
| F | F | T | T | F | F | F | 0.0625 | 2 | T | F | T | F | T | F | F | 0.03125 | 1 |
| F | F | T | T | F | F | T | 0.03125 | 1 | T | F | T | T | F | F | F | 0.03125 | 1 |
| F | T | F | F | F | F | F | 0.03125 | 1 | T | F | T | T | F | T | T | 0.0625 | 2 |
| F | T | F | T | F | F | F | 0.0625 | 2 | T | T | F | F | F | T | F | 0.03125 | 1 |
| F | T | F | T | F | F | T | 0.03125 | 1 | T | T | F | T | F | F | F | 0.03125 | 1 |
| F | T | T | F | F | F | F | 0.03125 | 1 | T | T | F | T | T | F | T | 0.0625 | 2 |
| F | T | T | T | F | F | F | 0.0625 | 2 | T | T | T | T | F | F | F | 0.125 | 4 |

To compute $E(\text{SD}, \text{OBS}|\alpha_3 \wedge \neg i)$, we have to find the sum and the sum of the squares of the intersection sizes of all rows in Table 2 for which column $i$ is **F**. It can be checked that $E(\text{SD}, \text{OBS}|\alpha_3, \neg i) = 24/16 = 1.5$. Similarly, $E(\text{SD}, \text{OBS}|\alpha_3 \wedge i) = 34/16 = 2.125$. Hence an optimal diagnostician would consider a second measurement with control setting $\gamma = i$.

The obvious problem with the above brute-force approach is that the size of the marginalization table is, in the worst-case, exponential in |OBS|. Although many of the rows in the marginalization table can be skipped as the intersections are empty (there are no consistent prior diagnoses with the respective observation vector and control assignment), the construction of this table is computationally so demanding that we will consider an approximation algorithm (to construct Table 1 for our tiny example, the exhaustive approach had to perform a total of 512 consistency checks).

### 5.2 Fractal$^{\text{ATPG}}$

Consider the running example from Sec. 3 and an observation $\alpha_4 = a \wedge b \wedge i \wedge o_1 \wedge \neg o_4$. This leads to the 6 double-fault MC diagnoses, shown in Fig. 3.

|  | $h_1$ | $h_2$ | $h_3$ | $h_4$ | $h_5$ | $h_6$ | $h_7$ | $h_8$ |
|---|---|---|---|---|---|---|---|---|
| $\omega_1$ | ✓ | ✗ | ✓ | ✗ | ✗ | ✗ | ✗ | ✗ |
| $\omega_2$ | ✓ | ✗ | ✗ | ✗ | ✓ | ✗ | ✗ | ✗ |
| $\omega_3$ | ✗ | ✓ | ✗ | ✗ | ✓ | ✗ | ✗ | ✗ |
| $\omega_4$ | ✗ | ✗ | ✓ | ✗ | ✓ | ✗ | ✗ | ✗ |
| $\omega_5$ | ✗ | ✗ | ✗ | ✓ | ✓ | ✗ | ✗ | ✗ |
| $\omega_6$ | ✗ | ✗ | ✗ | ✗ | ✓ | ✗ | ✗ | ✓ |
| $E^\leq$ | $\frac{10}{3}$ | $\frac{13}{3}$ | $\frac{10}{3}$ | $\frac{13}{3}$ | $\frac{13}{3}$ | 6 | 6 | $\frac{13}{3}$ |

Figure 3: ATPG-Based active testing example

Instead of searching through the space of all possible control assignments, we directly compute a control assignment that tests a specific component $c$ by using an approach from FELDMAN, PROVAN, & VAN GEMUNDATPG. We choose this component $c$ to be the one that most decreases the expected number of remaining MC diagnoses by minimizing $E^{\leq}(\text{SD}, c|\alpha \wedge \gamma)$. If we look at Fig. 3 we can see that knowing the health of $h_1$ and $h_3$ leads to $E^{\leq} \approx 3.33$, for $h_2$, $h_4$, $h_5$, and $h_7$, we have $E^{\leq} \approx 4.33$, and for $h_6$ and $h_7$ we have $E^{\leq} = 6$. Choosing a control setting that computes the state of $h_1$ or $h_3$ is intuitive as the state of this component makes the most balanced partition of the prior diagnoses.

We next present the FRACTAL$^{\text{ATPG}}$ algorithm that uses the approach illustrated above.

---

**Algorithm 2** ATPG-Based active testing algorithm

1: **function** FRACTAL$^{\text{ATPG}}$(ATS, $\alpha$, $\gamma$) **returns** a control term
    **inputs:** ATS (active testing system): model
             $\alpha$ (term): initial (non-modifiable) observation
             $\gamma$ (term): initial control
    **local variables:** $c$ (variable): component
                      $f$ (integer): remaining diagnoses
                      $d$ (term): diagnosis
                      $\gamma$ (term): control setting
                      $H$ (set of pairs): component/expectation pairs
                      $D$ (set of terms): diagnoses
2:     $D \leftarrow \Omega^{\leq}(\text{SD}, \alpha \wedge \gamma)$
3:     **for all** $c \in$ COMPS **do**
4:         $f \leftarrow 0$
5:         **for all** $d \in D$ **do**
6:             **if** $c \in d$ **then**
7:                 $f \leftarrow f + 1$
8:             **end if**
9:         **end for**
10:        $H \leftarrow \langle c, f^2 + (|D| - f)^2 \rangle$
11:    **end for**
12:    $H \leftarrow$ SORTBYEXPECTATION($H$)
13:    **for** $i = 1 \ldots |H|$ **do**
14:       **if** $\gamma \leftarrow$ ATPG(ATS, $\alpha$, $H_i\langle c \rangle$) **then**
15:          **return** $\gamma$
16:       **end if**
17:    **end for**
18:    **return** RANDOMCONTROLS()
19: **end function**

---

Algorithm 2 counts the number of prior diagnoses that each component appears in (lines 4 - 8) and the result is saved in the variable $f$. This number is then used to compute the expected number of remaining MC diagnoses given the component health (line 10). For each component the expected number of diagnoses is stored in the set $H$ (line 10). The set $H$ is then sorted in increasing order of expectation (line 12). We then iterate over the set of components in order of expectation (lines 13 – 17). For each component we try to compute

314

an ATPG vector that tests it. In some cases such a vector may not exist. In the worst case there is no ATPG vector that can test any component, and Alg. 2 has no better strategy but to return a random control assignment (line 18).

The time complexity of Alg. 2 is determined by the complexity of the diagnostic search (line 2) and the complexity of ATPG (line 14). If we denote the former with $\Psi$ and the latter with $\Phi$ then the complexity of FRACTAL$^{\text{ATPG}}$ becomes $O(\Phi\Psi|\text{COMPS}|)$. As the complexity of ATPG is usually lower than that of diagnosis (abductive reasoning) ($\Phi < \Psi$), the complexity of FRACTAL$^{\text{ATPG}}$ is determined by the time for computing MC diagnoses.

Computing ATPG vectors has been extensively studied (Bushnell & Agrawal, 2000) and although it is known to be an *NP*-hard problem (Ibarra & Sahni, 1975), there exists evidence that ATPG is easy for practical problems (Prasad, Chong, & Keutzer, 1999). Some efficient ATPG algorithms integrate randomized approach and Boolean difference (Bushnell & Agrawal, 2000). The former approach efficiently computes test vectors for the majority of components, while the latter computes test vectors for the remaining components by using a DPLL-solver.

We implement ATPG as follows. First we duplicate the system description SD by renaming each variable $v : v \notin \{\text{IN} \cup \text{CTL}\}$ to $v'$, thus generating SD$'$ (SD and SD$'$ share the same input and control variables). Then we create the all healthy assignment (for all assumable variables) $\omega^0$ and the single fault assignment $\omega^I$ such that $\omega^0$ and $\omega^I$ differ only in the sign of the literal whose component we want to test. Finally, we construct the following propositional expression:

$$\Delta \equiv \alpha \wedge \text{SD} \wedge \text{SD}' \wedge \omega^0 \wedge \omega^I \wedge \left[ \bigvee_{o \in \text{OUT}} o \oplus o' \right] \tag{7}$$

where the operator $\oplus$ denotes an exclusive or, hence $o \oplus o' \equiv (\neg o \wedge o') \vee (o \wedge \neg o')$.

The propositional expression in (7) leaves unconstrained only the controls $\gamma$ that we need. There are two "instances" of the system: healthy (SD and $\omega^0$) and faulty (SD and $\omega^I$). The last term in $\Delta$ forces the output of the healthy and the faulty system to be different in at least one bit. To compute an ATPG control vector we need one satisfiable solution of $\Delta$. Note that an ATPG control vector may not exist ($\Delta \models \bot$), i.e., a component may not be testable given CTL and SD $\wedge \alpha$. Often there are multiple satisfying control assignments. In this case FRACTAL$^{\text{ATPG}}$ chooses an arbitrary one. The latter does not mean that all satisfiable ATPG control vectors achieve the same uncertainty reduction. FRACTAL$^{\text{ATPG}}$ becomes suboptimal when there is no control testing a given component, or when there are multiple controls. FRACTAL$^{\text{ATPG}}$ becomes completely random when there are no components that can be tested with the given choice of controls.

There are two problems with FRACTAL$^{\text{ATPG}}$. First, FRACTAL$^{\text{ATPG}}$ assumes stationary inputs, i.e., FRACTAL$^{\text{ATPG}}$ ignores a source of uncertainty. The non-modifiable inputs, however, can only help in the decay process, hence FRACTAL$^{\text{ATPG}}$ is "conservative" in choosing the control assignments–a feature that leads to suboptimality. A bigger problem is that FRACTAL$^{\text{ATPG}}$ decreases the expected number of remaining MC diagnoses by computing the exact health of one component. Here, the problem is not that FRACTAL$^{\text{ATPG}}$ tests *one* component per step, but that it tries to compute a control assignment that computes the exact state of this component. An active testing algorithm can decrease the diagnostic uncertainty by computing a probability distribution function for the state of each component.

A natural extension of FRACTAL$^{\text{ATPG}}$ is an algorithm that computes the state of $k$ components simultaneously. The latter approach assumes that the system is $k$-component testable–an unrealistic assumption. In our experiments we have seen that systems are often even not single-component testable. Note that computing the exact states of components is not a requirement for decreasing the diagnostic uncertainty. Instead of computing the exact state of one or more components, the algorithm shown in the next section implicitly builds a probability density function for the health state of each component, and does not suffer from the problems of FRACTAL$^{\text{ATPG}}$.

### 5.3 Fractal$^{\text{G}}$

Consider SD from the example started in Sec. 3, input variables IN = $\{i\}$, control variables CTL = $\{a, b\}$, initial input values $\beta = i$, and an initial observation $\alpha_3 = \beta \wedge (\neg a \wedge \neg b) \wedge (\neg o_1 \wedge o_2 \wedge o_3 \wedge o_4)$. The initial observation $\alpha_3$ leads to 5 triple-fault MC diagnoses: $\Omega^{\leq}(\text{SD}, \alpha_3) = \{\{\neg h_1, \neg h_4, \neg h_7\}, \{\neg h_1, \neg h_7, \neg h_8\}, \{\neg h_2, \neg h_3, \neg h_6\}, \{\neg h_2, \neg h_4, \neg h_5\}, \{\neg h_3, \neg h_6, \neg h_8\}\}$. We also write $D = \Omega^{\leq}(\text{SD}, \alpha_3)$ and choose one of the faults in $D$ to be the truly injected fault $\omega^*$ (let $\omega^* = \{\neg h_1, \neg h_7, \neg h_8\}$).

| | Exhaustive | | | Greedy | |
|---|---|---|---|---|---|
| $k$ | $\gamma_1$ | $E^{\leq}(\text{SD}, \text{IN}\|\gamma_1)$ | $k$ | $\gamma_1$ | $E^{\leq}(\text{SD}, \text{IN}\|\gamma_1)$ |
| 1 | $\neg a \wedge \neg b$ | 4.33 | 1 | $\neg a \wedge \neg b$ | 4.33 |
| 2 | $a \wedge \neg b$ | 1.57 | 2 | $\mathbf{a} \wedge \neg b$ | 1.57 |
| 3 | $\neg a \wedge b$ | 1.57 | 3 | $a \wedge \mathbf{b}$ | 1.33 |
| 4 | $a \wedge b$ | 1.33 | | | |
| | $|\Omega^{\cap}(D, \beta \wedge \gamma_1)| = 2$ | | | $|\Omega^{\cap}(D, \beta \wedge \gamma_1)| = 2$ | |

| $k$ | $\gamma_2$ | $E^{\leq}(\text{SD}, \text{IN}\|\gamma_2)$ | $k$ | $\gamma_2$ | $E^{\leq}(\text{SD}, \text{IN}\|\gamma_2)$ |
|---|---|---|---|---|---|
| 1 | $\neg a \wedge \neg b$ | 1.67 | 1 | $\neg a \wedge \neg b$ | 1.67 |
| 2 | $a \wedge \neg b$ | 1 | 2 | $\neg a \wedge \mathbf{b}$ | 1 |
| 3 | $\neg a \wedge b$ | 1 | 3 | $a \wedge b$ | 1.67 |
| 4 | $a \wedge b$ | 1.67 | | | |
| | $|\Omega^{\cap}(\Omega^{\cap}(D, \beta \wedge \gamma_1), \beta \wedge \gamma_2)| = 1$ | | | $|\Omega^{\cap}(\Omega^{\cap}(D, \beta \wedge \gamma_1), \beta \wedge \gamma_2)| = 1$ | |

Figure 4: Exhaustive and greedy search for an optimal control assignment

The left and right parts of Fig. 4 show two possible scenarios for locating $\omega^*$. On the left we have an exhaustive approach which considers all the $2^{|\text{CTL}|}$ control assignments, hence it cannot be used to solve practical problems. The greedy scenario on the right side of Fig. 4 decreases the number of computations of expected number of remaining MC diagnoses from $2^{|\text{CTL}|}$ to $|\text{CTL}|$. The idea is to flip one control variable at a time, to compute the expected number of remaining MC diagnoses and to keep the flip (shown in bold in Fig. 4) if $E^{\leq}$ decreases. Given an initial control assignment $\gamma$ we consider the space of possible control flips. This space can be visualized as a lattice (Fig. 5 shows a small example). Figure 5 shows the expected number of MC diagnoses for each control assignment. Note that probing can be visualized in a similar way.

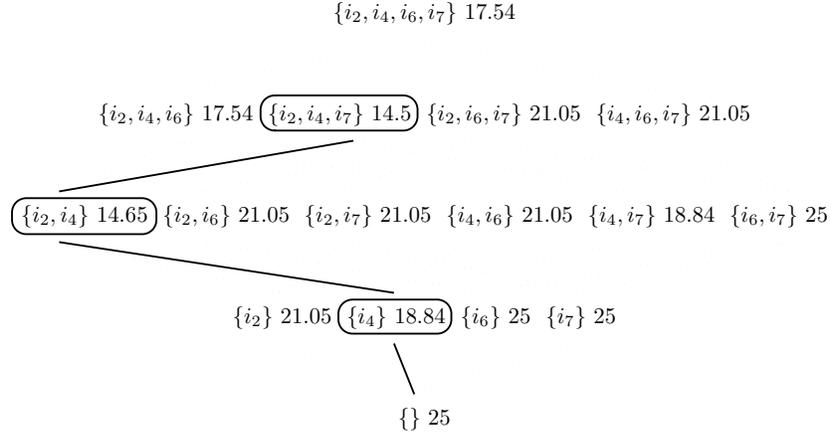

Figure 5: Example of an expectation optimization lattice $(74182, |CTL| = 4, |IN| = 5)$. Each node shows the set of control flips and the expected number of MC diagnoses.

In practice, control literals are mostly independent and even though the space of control assignments is not continuous in general, it has large continuous subspaces. The greedy approach is shown in Alg. 3, which computes a control assignment for a given active testing system and a prior observation.

---

**Algorithm 3** Greedy active testing algorithm
---
1: **function** FRACTAL(ATS, $\alpha$) **returns** a control term
    **inputs:** ATS (active testing system): model
        $\alpha$ (term): initial observation
    **local variables:** $\gamma, \gamma'$ (terms): control configurations
        $E, E'$ (reals): expectations
        $D$ (set of terms): diagnoses
        $l$ (literal): control literal
2:    $D \leftarrow \Omega^{\leq}(\text{SD}, \alpha)$
3:    $E \leftarrow \text{EXPECTATION}(\text{ATS}, \gamma, D)$
4:    **for all** $l \in \gamma$ **do**
5:        $\gamma' \leftarrow \text{FLIPLITERAL}(\gamma, l)$
6:        $E' \leftarrow \text{EXPECTATION}(\text{ATS}, \gamma', D)$
7:        **if** $E' < E$ **then**
8:            $\gamma \leftarrow \gamma'$
9:            $E \leftarrow E'$
10:       **end if**
11:    **end for**
12:    **return** $\gamma$
13: **end function**

The set of initial diagnoses is computed from the initial observation in line 2. In line 5, Alg. 3 "flips" the next literal in the current control assignment. The auxiliary FLIPLITERAL subroutine simply changes the sign of a specified literal in a term. After each "flip" the expected intersection size is computed with a call to EXPECTATION (cf. Alg. 1). If the new expected intersection size is smaller than the current one, then the proposed control assignment is accepted as the current control assignment, and the search continues from there.

The complexity of FRACTAL$^\text{G}$ is determined by the complexity of the diagnostic search (line 2) and the complexity of EXPECTATION (line 3 and line 6). If we denote the former with $\Psi$ and the latter with $\Xi$ then the complexity of FRACTAL$^\text{G}$ becomes $O(\Phi \Xi |\text{CTL}|)$. As $\Phi \sim \Xi$, the complexity of FRACTAL$^\text{G}$ is the same as FRACTAL$^\text{G}$. In practice FRACTAL$^\text{G}$ requires more computation to compute a sufficient decay. This is due to the design of EXPECTATION (Alg. 1).

While the active-testing problem is worst-case *NP*-hard (it can be reduced to computing a diagnosis), as we will see in the experimentation section, it is possible to achieve very good average-case performance by choosing an appropriate MBD oracle. The advantage of the greedy approach, in particular, is that the number of computations of the expected number of diagnoses is linear in the number of literals in the control assignment. This is done at the price of some optimality (i.e., the effect of combinations of controls is neglected).

### 5.4 Fractal$^\text{P}$

Probing is related to active testing as measuring internal variables can be thought of as revealing internal control circuits. Alternatively, one can add control circuitry to a model that reveals the values of internal variables. To reveal this hidden control potential we implement GDE probing (de Kleer & Williams, 1987) in FRACTAL$^\text{P}$. Our approach is different from GDE in two ways. First, we compute the expected number of remaining MC diagnoses instead of expected diagnostic entropy. Second, FRACTAL does not use an Assumption-Based Truth Maintenance System (ATMS) (de Kleer, 1986).

Consider the running example from Sec. 3 and an observation $\alpha_5 = \neg a \wedge \neg b \wedge i \wedge \neg o_1 \wedge o_2 \wedge o_3 \wedge o_4$. This leads to 5 triple-fault MC diagnoses: $\Omega^{\leq}(\text{SD}, \alpha_3) = \{\{\neg h_1, \neg h_4, \neg h_7\},$ $\{\neg h_1, \neg h_7, \neg h_8\}, \{\neg h_2, \neg h_3, \neg h_6\}, \{\neg h_2, \neg h_4, \neg h_5\}, \{\neg h_3, \neg h_6, \neg h_8\}\}$. Subsequent measurement of $p$ gives us $|\Omega^{\cap}(\Omega^{\leq}(\text{SD}, \alpha_3), p)| = 3$ if $p$ is positive and $|\Omega^{\cap}(\Omega^{\leq}(\text{SD}, \alpha_5), \neg p)| = 2$ otherwise. The expected number of MC diagnoses is $E^{\leq}(\text{SD}, \{p\}|\alpha_3) = 2.6$. Repeating this for the remaining internal variables results in $E^{\leq}(\text{SD}, \{q\}|\alpha_3) = 2.6$, $E^{\leq}(\text{SD}, \{r\}|\alpha_3) = 3.4$, and $E^{\leq}(\text{SD}, \{s\}|\alpha_3) = 3.4$. As a result we can see that measuring $p$ and $q$ is less informative than measuring $r$ and $s$, which is intuitive as $r$ and $s$ give a more balanced partitioning of the circuit.

**Problem 2** (Probe Sequencing). Given a system ATS, an observation $\alpha$ and a partial assignment to the internal variables $\psi$, choose a variable $p^*$ from the set $U$ of unassigned internal variables $\psi$, such that:

$$p^* = \underset{p \in U}{\arg\min}\, E^{\leq}(\text{SD}, p | \alpha \wedge \psi) \tag{8}$$

Algorithm 4 solves Problem 2. Algorithm 4 computes the expected number of diagnoses for each unobserved variable (lines 3 - 11). Starting from the set $D$ of initial diagnoses (computed in line 2), Alg. 4 perform a total of $2|D||V \setminus \{\text{OBS} \cup \text{COMPS}\}|$ consistency checks (lines 4 and 5) to determine the expected number of MC diagnoses for each unobserved variable.

We next show the probing algorithm as introduced by de Kleer and Williams (1987) and adapted for the Fractal framework.

---

**Algorithm 4** Probing algorithm
---
1: **function** Fractal$^P$(ATS, $\alpha$) **returns** a variable
    **inputs:** ATS (active testing system): model
              $\alpha$ (term): observation
    **local variables:** $v, R$ (variables): probes
                      $E, E'$ (reals): expectations
                      $p, q$ (reals): remaining diagnoses
                      $D$ (set of terms): diagnoses
2:    $D \leftarrow \Omega^{\leq}(\text{SD}, \alpha)$
3:    **for all** $v \in V \setminus \{\text{COMPS} \cup \text{OBS}\}$ **do**
4:        $p \leftarrow |\Omega^{\cap}(D, \alpha \wedge v)|$
5:        $q \leftarrow |\Omega^{\cap}(D, \alpha \wedge \neg v)|$
6:        $E' \leftarrow (p^2 + q^2)/(p + q)$
7:        **if** $E' < E$ **then**
8:            $R \leftarrow v$
9:            $E \leftarrow E'$
10:       **end if**
11:   **end for**
12:   **return** $R$
13: **end function**

---

Instead of computing the expected number of remaining MC diagnoses for a single variable $p$, it is possible to consider measuring all pairs of variables $\langle p_1, p_2 \rangle$, or in general, all $k$-tuples of internal variables $\langle p_1, p_2, \ldots, p_m \rangle$ for $m \leq |V \setminus \{\text{OBS} \cup \text{COMPS}\}|$. We will refer to probing involving more than 1 variable as $k$-probing. Although it has been shown that users do not significantly benefit in terms of diagnostic uncertainty by performing $k$-probing (de Kleer et al., 1992), we can easily modify Fractal$^P$ to consider multiple probes. Note that for $m = |V \setminus \{\text{OBS} \cup \text{COMPS}\}|$ there is no probing problem, as there is only one way to pick all internal variables.

The most complex operation in Fractal$^P$ is again computing the initial set of MC diagnoses. In addition to that, we have $|V \setminus \{\text{COMPS} \cup \text{OBS}\}|$ consistency checks. Consistency checking is, in general, easier than diagnosis. Note that all Fractal algorithms (Fractal$^{\text{ATPG}}$, Fractal$^G$, and Fractal$^P$) start with computing the set of initial MC diagnoses. Hence, the difference in their performance is determined by the complexity of reducing the initial set $\Omega^{\leq}(\text{SD}, \alpha)$. According to this criterion, the fastest algorithm is Fractal$^P$ as it only performs a small number of consistency checks, followed closely by Fractal$^{\text{ATPG}}$ (computing ATPG vectors). The slowest algorithm is Fractal$^G$, as it computes the expected number of MC diagnoses given multiple variables.

## 6. Experimental Results

We have implemented FRACTAL in approximately 3 000 lines of C code (excluding the diagnostic engine and the Logic Based Truth Maintenance System). All experiments were run on a 64-node dual-CPU cluster (each node configured with two 2.4 GHz AMD Opteron DP 250 processors and 4 Gb of RAM).

### 6.1 Experimental Setup

We have experimented on the well-known benchmark models of ISCAS85 combinational circuits (Brglez & Fujiwara, 1985). As models derived from the ISCAS85 circuits are computationally intensive (from a diagnostic perspective), we have also considered four medium-sized circuits from the 74XXX family (Hansen, Yalcin, & Hayes, 1999). In order to use the same system model for *both* MC diagnosis counting and simulation, the fault mode of each logic gate is "stuck-at-opposite", i.e., when faulty, the output of a logic gate assumes the opposite value from the nominal. Without loss of generality, only gates are allowed to fail in our models. This is different from ATPG where gates typically do not fail but wires are modeled as components that can fail with failure modes "stuck-at-zero" and "stuck-at-one". The ATPG and MBD modeling approaches achieve the same results.

Table 3: An overview of the 74XXX/ISCAS85 circuits ($V$ is the total number of variables and $C$ is the number of clauses)

| Name | Description | |IN| | |OUT| | Original |COMPS| | $V$ | $C$ | Reduced |COMPS| |
|---|---|---|---|---|---|---|---|
| 74182 | 4-bit CLA | 9 | 5 | 19 | 47 | 150 | 6 |
| 74L85 | 4-bit comparator | 11 | 3 | 33 | 77 | 236 | 15 |
| 74283 | 4-bit adder | 9 | 5 | 36 | 81 | 244 | 14 |
| 74181 | 4-bit ALU | 14 | 8 | 65 | 144 | 456 | 21 |
| c432 | 27-channel interrupt ctl. | 36 | 7 | 160 | 356 | 514 | 59 |
| c499 | 32-bit SEC circuit | 41 | 32 | 202 | 445 | 714 | 58 |
| c880 | 8-bit ALU | 60 | 26 | 383 | 826 | 1 112 | 77 |
| c1355 | 32-bit SEC circuit | 41 | 32 | 546 | 1 133 | 1 610 | 58 |
| c1908 | 16-bit SEC/DEC | 33 | 25 | 880 | 1 793 | 2 378 | 160 |
| c2670 | 12-bit ALU | 233 | 140 | 1 193 | 2 695 | 3 269 | 167 |
| c3540 | 8-bit ALU | 50 | 22 | 1 669 | 3 388 | 4 608 | 353 |
| c5315 | 9-bit ALU | 178 | 123 | 2 307 | 4 792 | 6 693 | 385 |
| c6288 | 32-bit multiplier | 32 | 32 | 2 416 | 4 864 | 7 216 | 1 456 |
| c7552 | 32-bit adder | 207 | 108 | 3 512 | 7 232 | 9 656 | 545 |

In addition to the original 74XXX/ISCAS85 models, we have performed cone reductions as described by Siddiqi and Huang (2007) and de Kleer (2008). Recall that from the perspective of the MBD diagnostic engine, faults *inside* a cone (where a cone is a set of components) cannot be distinguished, hence it is enough to provide a single health variable per cone. We call models with a single health variable per cone "reduced". Table 3 describes all models.

Both initial observation vectors and control settings are used in the first step of the Fractal inference. To illustrate the significant diagnostic convergence that is possible, we use initial observations leading to high numbers of initial MC diagnoses.

To average over the diagnostic outcomes of the observations, we repeat each experiment with a range of initial observation vectors. The cardinality of the MC diagnosis is of no significance to Fractal, but it produces a significant burden on the diagnostic oracle (Feldman, Provan, & van Gemund, 2008). In order to overcome this computational difficulty, we have limited our experiments to observation vectors leading to double faults only.

For each circuit we have generated 1 000 non-masking double-faults, and for each observation we have computed the number of initial MC diagnoses. From the 1 000 observation vectors we have taken the 100 with the largest number of MC diagnoses. The resulting observations are summarized in Table 4. For example, we can see a staggering number of 46 003 double faults for the most under-constrained c7552 observation.

Table 4: Number of MC diagnoses per observation vector

|  | Original | | | Reduced | | |
| --- | --- | --- | --- | --- | --- | --- |
| Name | Min | Max | Mean | Min | Max | Mean |
| 74182 | 25 | 25 | 25 | 2 | 2 | 2 |
| 74L85 | 32 | 88 | 50.2 | 5 | 13 | 7 |
| 74283 | 48 | 60 | 51.8 | 6 | 9 | 7 |
| 74181 | 93 | 175 | 113.5 | 10 | 19 | 13.3 |
| c432 | 88 | 370 | 165.8 | 21 | 127 | 47.5 |
| c499 | 168 | 292 | 214.1 | 4 | 31 | 15.8 |
| c880 | 783 | 1 944 | 1 032.5 | 20 | 190 | 39.7 |
| c1355 | 1 200 | 1 996 | 1 623.3 | 4 | 31 | 15 |
| c1908 | 1 782 | 5 614 | 2 321.9 | 40 | 341 | 84.8 |
| c2670 | 2 747 | 7 275 | 3 621.7 | 10 | 90 | 21.3 |
| c3540 | 1 364 | 2 650 | 1 642.2 | 158 | 576 | 226 |
| c5315 | 3 312 | 17 423 | 6 202.1 | 15 | 192 | 34.5 |
| c6288 | 6 246 | 15 795 | 8 526.1 | 2 928 | 6 811 | 3 853 |
| c7552 | 16 617 | 46 003 | 23 641.2 | 45 | 624 | 121.6 |

Since the 74XXX/ISCAS85 circuits have no control variables we "abuse" the benchmark by designating a fraction of the input variables as controls.

We define two policies for generating next inputs: *random* and *stationary*. The latter input policy (where the input values do not change in time) is a typical diagnostic worst-case for system environments which are, for example, paused pending diagnostic investigation, and it provides us with useful bounds for analyzing Fractal's performance.

Note that the use of non-characterizing sets of diagnoses (see Def. 4) may lead to a situation in which the real (injected) fault is not in the initial set of diagnoses. In such a case the set of remaining diagnoses $\Omega(S)$ may become an empty set after some number of Fractal steps. Although this gives us some diagnostic information, this is an undesirable situation and non-characterizing sets of diagnoses should represent most of the diagnostic probability mass to minimize the likelihood of such cases. We have constructed our experimental benchmark of initial observations in such a way as to avoid such cases.

### 6.2 Expected Number of MC Diagnoses

We have observed that the error of Alg. 1 is insensitive to the number or the composition of the input variables. It can be seen that the value of the expected number of diagnoses $\hat{E}$ approaches the exact value $E$ when increasing the number of samples $n$. In particular, $\hat{E}$ is equal to the exact value of the expected number of MC diagnoses $E$, when all possible input values are considered. Figure 6 shows examples of $\hat{E}$ approaching $E$ for three of our benchmark models.

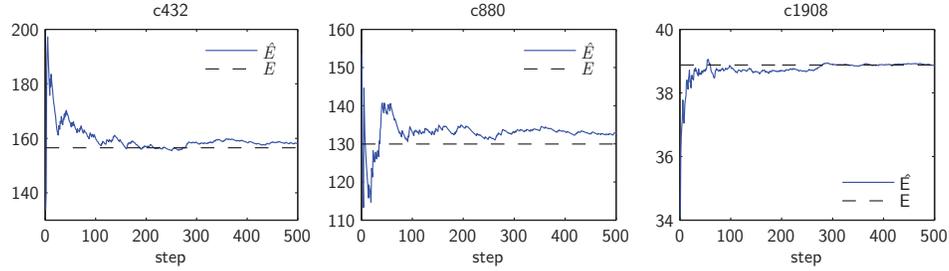

Figure 6: Convergence of expected number of MC diagnoses with increasing sample size

TERMINATE approximates the intermediate value of $\hat{E}$ by computing the sequence $\mathbf{E} = \langle \hat{E}_1, \hat{E}_2, \ldots, \hat{E}_n \rangle$. The standard error of the mean of $\mathbf{E}$ is defined as:

$$\text{SEM}_\mathbf{E} = \frac{s}{\sqrt{n}}, \qquad (9)$$

where $s$ is the standard deviation of $\mathbf{E}$. We have set TERMINATE to terminate Alg. 1 when $n > 15$ and $\text{SEM}_\mathbf{E} < \theta$, where $\theta$ is a circuit-dependent threshold constant. Table 5 shows $\theta$ for the various circuits we have experimented on.

Table 5: Termination parameters for Alg. 1

| Name | Original $\theta$ | Mean $n$ | Max $n$ | Reduced $\theta$ | Mean $n$ | Max $n$ |
|---|---|---|---|---|---|---|
| 74182 | 0.1 | 52.7 | 110 | 0.01 | 151.6 | 223 |
| 74L85 | 0.11 | 176.2 | 246 | 0.03 | 170.3 | 212 |
| 74283 | 0.2 | 139.9 | 225 | 0.01 | 211.3 | 243 |
| 74181 | 0.4 | 169.1 | 203 | 0.07 | 143.2 | 181 |
| c432 | 0.72 | 48.2 | 99 | 0.18 | 108.6 | 158 |
| c499 | 0.77 | 36.4 | 61 | 0.02 | 55.7 | 92 |
| c880 | 3.57 | 93.9 | 163 | 0.1 | 156.3 | 204 |
| c1355 | 4.3 | 51.3 | 93 | 0.01 | 121.4 | 148 |
| c1908 | 14.01 | 19.5 | 35 | 0.62 | 18.8 | 25 |
| c2670 | 12.77 | 40.5 | 78 | 0.1 | 65.2 | 102 |
| c3540 | 13.6 | 78.9 | 196 | 0.66 | 89.6 | 132 |
| c5315 | 23.35 | 34.2 | 39 | 0.09 | 36.0 | 48 |
| c6288 | 33.18 | 37.2 | 144 | 19.1 | 39.4 | 74 |
| c7552 | 68.11 | 68.7 | 91 | 3.73 | 73 | 122 |

We have determined $\theta$ using the following procedure. First, for each circuit, we choose an arbitrary initial observation and a small set IN of input variables ($|\text{IN}| = 8$). The small cardinality of IN allows us to compute true values of $E$. Next, for each circuit we run 10 pseudo-random experiments. For $\theta$ we choose the smallest value of $\text{SEM}_\mathbf{E}$ such that its corresponding $\hat{E}$ is within 95% of $E$. Table 5 shows the average and maximum number of steps in which Alg. 1 reaches this value. In all cases an upper bound of $n = 100$ is a safe termination criterion.

### 6.3 Comparison of Algorithms

Consider a weak-fault model of a chain of $n$ inverters and a set of MC diagnoses $D$ (initially, $|D| = n$). At each step single-variable probing can eliminate at most $0.5|D|$ diagnoses. It can also be shown that halving the expected number of remaining MC diagnoses is a theoretical bound of any one-step lookahead strategy. As a result we use the geometric decay curve

$$N(k) = N_0 p^k + N_\infty \tag{10}$$

as a model of the diagnosis decay. In this case, $N_0$ is the initial number of diagnoses, $N_\infty$ is the value to which $|\Omega(S)|$ converges, and $p$ is the decay rate constant. For probing, $N_\infty = 1$. In all our experiments we will fit both the expected number of remaining MC diagnoses $E$ and the actual number or remaining MC diagnoses $\Omega(S)$ to Eqn. 10.

#### 6.3.1 Fractal[ATPG]

Figure 7 shows the reduction of the expected number of MC diagnoses as a function of (1) the number of control variables $|\text{CTL}|$ and (2) the time $k$. One can easily see that a global optimum is reached quickly on both independent axes. This decay is shown for both c432 (Fig. 7, left) and the reduced c880 (Fig. 7, right). The number of control variables $|\text{CTL}|$ varies from 0 to 36 for c432 ($|\text{IN}| = 36$) and from 0 to 60 for c880 ($|\text{IN}| = 60$).

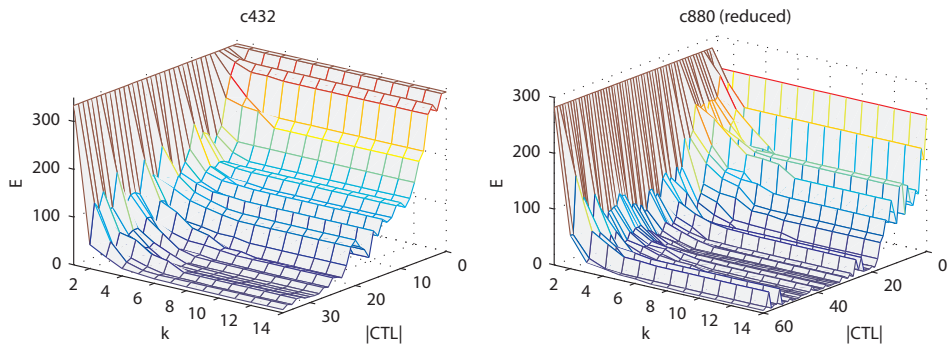

Figure 7: Decay of $E$, stationary inputs, Fractal[ATPG]

Using $|\Omega(S)|$ instead of $E$ results in similar plots (there is high correlation between $E$ and $|\Omega(S)|$), hence we have omitted the $|\Omega(S)|$ plots. The minimum, maximum and mean Pearson's linear correlation coefficient between $E$ from Fig. 7 and the respective $|\Omega(S)|$ for each number of control variables in c432 is $\rho_{\min} = 0.713$, $\rho_{\max} = 0.999$, and $\rho_{\text{avg}} = 0.951$, respectively. The corresponding correlation coefficients for the reduced c880 are $\rho_{\min} = 0.834$, $\rho_{\max} = 1$, and $\rho_{\text{avg}} = 0.972$.

It can be seen that the expected number of remaining diagnoses $E$ quickly reaches a global optimum when increasing $|\text{CTL}|$, which means that turning even a small number of input variables into controls allows for a geometric decay of the diagnostic entropy. The results for the reduced c880 are similar to the non-reduced c432. Hence, identification of cones helps the performance of the diagnostic oracle, but does not change the convergence behavior or the effect of the control variables.

Fitting geometric decay curves (Eqn. 10) on the $|\text{CTL}|$ axes of Fig. 7 produces better fits for c880 than for c432. Similarly, the values of $N_\infty$ for fits alongside the $k$-axis are larger for c432 than for c880. The reason for that is the small number of outputs in c432 (cf. Table 3). In circuits with few outputs, randomly turning a limited number of inputs into controls may not lead to a fast decay or a small $N_\infty$, as the control-output connectivity of a model is essential for decreasing the diagnostic uncertainty.

Table 6 and Table 7 summarize a total of $14\,000$ FRACTAL$^{\text{ATPG}}$ experiments over the whole 74XXX/ISCAS85 benchmark. Table 6 shows the correlation between the expected number of remaining MC diagnoses and the actual number of remaining MC diagnoses. In the second and third columns of Table 6 we can see the minimum and average correlations between $E$ and $\Omega^\leq(S)$. The third and fourth cases specify the fraction of observations for which we have $\rho > 0.95$ and $\rho > 0.975$, respectively. Columns 6 – 9 repeat this data for the reduced 74XXX/ISCAS85 circuits.

Table 6: Linear correlation coefficient $\rho$ of the expected number of remaining MC diagnoses $E$ and the actual number of remaining diagnoses $\Omega^\leq(S)$, $|\text{CTL}| = \frac{1}{4}|\text{IN}|$, stationary inputs, FRACTAL$^{\text{ATPG}}$

| | Original | | | | Reduced | | | |
|---|---|---|---|---|---|---|---|---|
| Name | $\rho_{\min}$ | $\rho_{\text{avg}}$ | $\rho > 0.95$ | $\rho > 0.975$ | $\rho_{\min}$ | $\rho_{\text{avg}}$ | $\rho > 0.95$ | $\rho > 0.975$ |
| 74182 | 0.55 | 0.98 | 0.82 | 0.79 | 1 | 1 | 1 | 1 |
| 74L85 | 0.46 | 0.91 | 0.52 | 0.44 | 0.45 | 0.81 | 0.4 | 0.39 |
| 74283 | 0.46 | 0.91 | 0.69 | 0.61 | 0.45 | 0.84 | 0.38 | 0.31 |
| 74181 | 0.46 | 0.88 | 0.48 | 0.39 | 0.45 | 0.86 | 0.44 | 0.35 |
| c432 | 0 | 0.83 | 0.24 | 0.16 | 0 | 0.81 | 0.29 | 0.23 |
| c499 | 0.5 | 0.87 | 0.32 | 0.15 | 0 | 0.86 | 0.42 | 0.33 |
| c880 | 0.51 | 0.86 | 0.28 | 0.18 | 0.51 | 0.85 | 0.3 | 0.19 |
| c1355 | 0 | 0.88 | 0.32 | 0.18 | 0 | 0.87 | 0.47 | 0.34 |
| c1908 | 0.38 | 0.87 | 0.31 | 0.18 | 0 | 0.79 | 0.22 | 0.15 |
| c2670 | 0 | 0.89 | 0.31 | 0.16 | 0 | 0.79 | 0.19 | 0.12 |
| c3540 | 0.48 | 0.86 | 0.29 | 0.19 | 0.06 | 0.82 | 0.3 | 0.23 |
| c5315 | 0.54 | 0.93 | 0.48 | 0.39 | 0.47 | 0.88 | 0.44 | 0.32 |
| c6288 | 0.42 | 0.9 | 0.41 | 0.24 | 0.4 | 0.9 | 0.36 | 0.21 |
| c7552 | 0.62 | 0.88 | 0.44 | 0.13 | 0.45 | 0.89 | 0.43 | 0.23 |

Table 7 summarizes the parameters of the geometric decay curves fitted to $\Omega(S)$. We can see that although $\Omega(S)$ is well approximated by a geometric decay curve (the average goodness-of-fit criterion $R^2$ is 0.84) the average decay constant $p$ is low (0.13 for the non-reduced and 0.22 for the reduced 74XXX/ISCAS85 circuits).

Table 7: Decay rate $p$ (minimal, maximal, and average) and goodness-of-fit $R^2$ (average) of geometric decay best-fit to $\Omega(S)$, $|\text{CTL}| = \frac{1}{4}|\text{IN}|$, stationary inputs, FRACTAL$^{\text{ATPG}}$

| Name | Original | | | | Reduced | | | |
|---|---|---|---|---|---|---|---|---|
| | $p_{\min}$ | $p_{\max}$ | $p_{\text{avg}}$ | $R^2_{\text{avg}}$ | $p_{\min}$ | $p_{\max}$ | $p_{\text{avg}}$ | $R^2_{\text{avg}}$ |
| 74182 | 0.3 | 0.52 | 0.43 | 0.95 | 0.5 | 0.5 | 0.5 | 1 |
| 74L85 | 0.06 | 0.75 | 0.48 | 0.88 | 0.35 | 0.64 | 0.5 | 0.92 |
| 74283 | 0.18 | 0.68 | 0.57 | 0.78 | 0.31 | 0.6 | 0.48 | 0.94 |
| 74181 | 0.11 | 0.71 | 0.5 | 0.86 | 0.18 | 0.64 | 0.5 | 0.9 |
| c432 | 0.03 | 0.8 | 0.56 | 0.81 | 0.03 | 0.79 | 0.52 | 0.82 |
| c499 | 0.1 | 0.79 | 0.64 | 0.84 | 0.48 | 0.76 | 0.65 | 0.84 |
| c880 | 0.06 | 0.84 | 0.54 | 0.83 | 0.06 | 0.8 | 0.53 | 0.87 |
| c1355 | 0.03 | 0.9 | 0.62 | 0.81 | 0.39 | 0.76 | 0.63 | 0.85 |
| c1908 | 0.02 | 0.93 | 0.65 | 0.68 | 0.51 | 0.74 | 0.64 | 0.77 |
| c2670 | 0.05 | 0.91 | 0.63 | 0.77 | 0.14 | 0.75 | 0.6 | 0.8 |
| c3540 | 0.02 | 0.87 | 0.52 | 0.85 | 0.04 | 0.76 | 0.45 | 0.89 |
| c5315 | 0.22 | 0.91 | 0.65 | 0.8 | 0.06 | 0.79 | 0.54 | 0.84 |
| c6288 | 0.02 | 0.91 | 0.52 | 0.88 | 0.02 | 0.9 | 0.51 | 0.89 |
| c7552 | 0.56 | 0.95 | 0.76 | 0.61 | 0.01 | 0.88 | 0.58 | 0.85 |
| Average | 0.13 | 0.82 | 0.58 | 0.81 | 0.22 | 0.74 | 0.55 | 0.87 |

The decay rate $p$ depends mostly on the circuit topology, hence the large variance in Table 7. Consider, for example, an artificial topology, where there are $n$ components, and $n$ output variables that produce the health-state of each component for a specific control assignment (e.g., a self-test). In this topology $p$ would be very small as a diagnostician needs at most one test (control assignment) to decrease the number of MC diagnoses to one.

The performance of FRACTAL$^{\text{ATPG}}$ is determined by the size of the model and the diagnostic oracle. In the above experiments the overall time for executing a single scenario varied from 3.4 s for 74182 to 1 015 s for c6288. The satisfiability problems in the ATPG part were always easy and the DPLL solver spent milliseconds in computing control assignments.

The decay rate of FRACTAL$^{\text{ATPG}}$ depends on the number and composition of controls. In what follows we will see that FRACTAL$^{\text{G}}$ can achieve a similar decay rate with a smaller number of control variables.

### 6.3.2 FRACTAL$^{\text{G}}$

Figure 8 shows the decay in the expected number of remaining MC diagnoses for FRACTAL$^{\text{G}}$. While the reduction is similar for c432, we can see a steeper reduction in the number of remaining MC diagnoses on both independent axes. Hence, the greedy algorithm is better than FRACTAL$^{\text{ATPG}}$ in identifying control combinations of small size, thereby leading to a better decay rate.

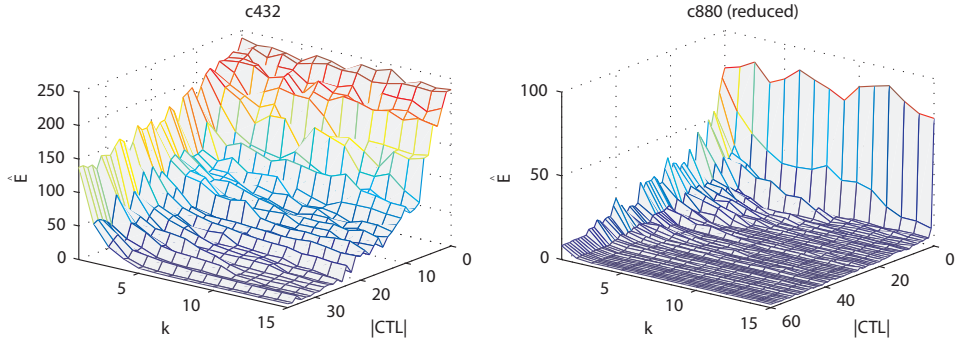

Figure 8: Decay of $E$ (left) and $\Omega(S)$ (right), stationary inputs, FRACTAL$^G$

Table 8 and Table 9 summarize the whole 74XXX/ISCAS85 benchmark. Table 8 shows that FRACTAL$^G$, similar to FRACTAL$^{ATPG}$, results in high average correlation between $\Omega(S)$ and $\hat{E}$ ($\rho_{\text{avg}} > 0.79$ for all circuits).

Table 8: Linear correlation coefficient $\rho$ of the expected number of remaining MC diagnoses $E$ and the actual number of remaining diagnoses $\Omega^{\leq}(S)$, $|\text{CTL}| = \frac{1}{4}|\text{IN}|$, stationary inputs, FRACTAL$^G$

|       | Original |       |             |              | Reduced |       |             |              |
|-------|----------|-------|-------------|--------------|---------|-------|-------------|--------------|
| Name  | $\rho_{\min}$ | $\rho_{\text{avg}}$ | $\rho > 0.95$ | $\rho > 0.975$ | $\rho_{\min}$ | $\rho_{\text{avg}}$ | $\rho > 0.95$ | $\rho > 0.975$ |
| 74182 | 0.03 | 0.88 | 0.39 | 0.18 | 1    | 1    | 1    | 1    |
| 74L85 | 0    | 0.72 | 0.12 | 0.06 | 0.01 | 0.66 | 0.16 | 0.14 |
| 74283 | 0    | 0.5  | 0.08 | 0.03 | 0    | 0.48 | 0.12 | 0.11 |
| 74181 | 0    | 0.56 | 0.05 | 0.02 | 0    | 0.55 | 0.09 | 0.07 |
| c432  | 0.01 | 0.75 | 0.07 | 0.02 | 0.01 | 0.68 | 0.07 | 0.05 |
| c499  | 0.01 | 0.88 | 0.29 | 0.08 | 0    | 0.85 | 0.33 | 0.2  |
| c880  | 0.05 | 0.77 | 0.09 | 0.06 | 0    | 0.73 | 0.08 | 0.04 |
| c1355 | 0.08 | 0.86 | 0.36 | 0.21 | 0.42 | 0.9  | 0.39 | 0.16 |
| c1908 | 0.05 | 0.81 | 0.25 | 0.14 | 0    | 0.8  | 0.4  | 0.3  |
| c2670 | 0.01 | 0.83 | 0.38 | 0.22 | 0.01 | 0.76 | 0.37 | 0.26 |
| c3540 | 0.34 | 0.73 | 0.09 | 0.05 | 0    | 0.7  | 0.04 | 0.01 |
| c5315 | 0.27 | 0.78 | 0.05 | 0    | 0    | 0.6  | 0.1  | 0.07 |
| c6288 | 0.09 | 0.81 | 0.11 | 0.05 | 0.1  | 0.78 | 0.09 | 0.04 |
| c7552 | 0.78 | 0.86 | 0.06 | 0.06 | 0.21 | 0.83 | 0.13 | 0.01 |

The decay rates of FRACTAL$^{ATPG}$ and FRACTAL$^G$ are similar (cf. Table 7 and Table 9), but, as is visible from Fig. 8, FRACTAL$^G$ reduces the number of remaining MC diagnoses more quickly, with fewer control variables. The c432 combinational circuit is difficult for active testing because it has a small number of outputs compared to the number of inputs (cf. Table 3), hence reducing the diagnostic utility.

To summarize the effect of the number of controls on the diagnostic convergence, we again fit the geometric decay curve (Eqn. 10) to $\Omega(S)$ for each of the 100 initial observation

A Model-Based Active Testing Approach to Sequential DiagnosisTable 9: Decay rate $p$ (minimal, maximal, and average) and goodness-of-fit $R^2$ (average) of geometric decay best-fit to $\Omega(S)$, $|\text{CTL}| = \frac{1}{4}|\text{IN}|$, stationary inputs, Fractal$^G$

|  | Original |  |  |  | Reduced |  |  |  |
|---|---|---|---|---|---|---|---|---|
| Name | $p_{\min}$ | $p_{\max}$ | $p_{\text{avg}}$ | $R^2_{\text{avg}}$ | $p_{\min}$ | $p_{\max}$ | $p_{\text{avg}}$ | $R^2_{\text{avg}}$ |
| 74182 | 0.24 | 0.53 | 0.43 | 0.95 | 0.5 | 0.5 | 0.5 | 1 |
| 74L85 | 0.05 | 0.74 | 0.47 | 0.9 | 0.25 | 0.65 | 0.49 | 0.93 |
| 74283 | 0.12 | 0.67 | 0.42 | 0.9 | 0.35 | 0.58 | 0.44 | 0.96 |
| 74181 | 0.15 | 0.75 | 0.48 | 0.9 | 0.15 | 0.69 | 0.44 | 0.93 |
| c432 | 0.04 | 0.88 | 0.56 | 0.83 | 0.03 | 0.86 | 0.59 | 0.8 |
| c499 | 0.09 | 0.88 | 0.71 | 0.81 | 0.34 | 0.85 | 0.68 | 0.85 |
| c880 | 0.12 | 0.67 | 0.42 | 0.9 | 0.07 | 0.83 | 0.52 | 0.88 |
| c1355 | 0.19 | 0.87 | 0.63 | 0.87 | 0.11 | 0.82 | 0.68 | 0.85 |
| c1908 | 0.32 | 0.73 | 0.53 | 0.87 | 0.05 | 0.84 | 0.59 | 0.83 |
| c2670 | 0.21 | 0.74 | 0.53 | 0.87 | 0.15 | 0.81 | 0.6 | 0.8 |
| c3540 | 0.34 | 0.63 | 0.53 | 0.91 | 0.01 | 0.8 | 0.44 | 0.9 |
| c5315 | 0.3 | 0.83 | 0.61 | 0.83 | 0.06 | 0.86 | 0.58 | 0.82 |
| c6288 | 0.04 | 0.81 | 0.5 | 0.9 | 0.08 | 0.77 | 0.47 | 0.89 |
| c7552 | 0.08 | 0.54 | 0.34 | 0.92 | 0.16 | 0.83 | 0.59 | 0.84 |
| Average | 0.16 | 0.73 | 0.51 | 0.88 | 0.17 | 0.76 | 0.54 | 0.88 |

vectors and various |CTL|. In this case, $N_0$ is the initial number of diagnoses, $N_\infty$ is the value to which $|\Omega(S)|$ converges, and $p$ is the decay constant (the most important parameter of our fits). For an "easy" circuit with chain topology, for $p = \frac{1}{2}$, $N_0$ halves every $k$ steps, as in binary search, hence $p$ corresponds to one bit. For $p = \frac{1}{4}$, $p$ corresponds to two bits.

Table 10: Mean $p$ for various numbers of control bits, stationary input policy, Fractal$^G$

|  | Original |  |  | Reduced |  |  |
|---|---|---|---|---|---|---|
| Name | 3 bits | 4 bits | 5 bits | 3 bits | 4 bits | 5 bits |
| c432 | 0.61 | 0.69 | 0.42 | 0.7 | 0.71 | 0.57 |
| c499 | 0.79 | 0.83 | 0.77 | 0.58 | 0.62 | 0.52 |
| c880 | 0.5 | 0.55 | 0.62 | 0.49 | 0.47 | 0.44 |
| c1355 | 0.71 | 0.72 | 0.59 | 0.8 | 0.82 | 0.75 |
| c1908 | 0.68 | 0.7 | 0.41 | 0.54 | 0.52 | 0.3 |
| c2670 | 0.45 | 0.49 | 0.39 | 0.39 | 0.44 | 0.42 |
| c3540 | 0.39 | 0.38 | 0.43 | 0.79 | 0.8 | 0.61 |
| c5315 | 0.52 | 0.62 | 0.67 | 0.81 | 0.72 | 0.79 |
| c6288 | 0.31 | 0.41 | 0.23 | 0.64 | 0.7 | 0.59 |
| c7552 | 0.62 | 0.77 | 0.3 | 0.59 | 0.34 | 0.38 |

Table 10 shows the average $p$ over all initial observations and for various numbers of control bits $b = \lg |\text{CTL}|$. Table 10 does not include data for the 74XXX circuits as they do not have enough inputs (we need circuits with at least 32 inputs). From Table 10 it is visible

327

Feldman, Provan, & van Gemund

that an exponential increase in the number of control variables does not lead to a significant decrease in $p$. Hence, for ISCAS85, even turning a small number of the input variables into controls leads to a near-optimal decrease in the number of remaining MC diagnoses.

The performance of FRACTAL$^G$ was worse than that of FRACTAL$^{ATPG}$ due to the multi-variable expectation. The running time varied between 7.1 s for 74182 and 2 382 s for c6288. Most of the CPU time was spent in the EXPECTATION subroutine (cf. Alg. 1). Each consistency check was computationally easy, but for each circuit there were thousands of them. Hence, improving the performance of LTMS would lead to an increase of the performance of FRACTAL$^G$.

### 6.3.3 FRACTAL$^P$

We next discuss FRACTAL$^P$. As mentioned earlier, probing is different from active testing as it assumes full observability of the model, i.e., all internal variables can be measured (cf. Sec. 5). Furthermore, probing considers one internal variable per step, while active testing assigns value to all control variables.[5]

The value of the decay rate $p$ depends on (1) the topology of the circuit, (2) the initial observation and (3) the values of the subsequent probes. For probing in ISCAS85 we see that the values of the decay rate $p$ are close to 0.5 for both $\Omega(S)$ and $E$. Figure 9 shows the actual and expected number of remaining MC diagnoses ($\Omega^{\leq}(S)$ and $E$, respectively) and a geometric fit to $E$ for three probing scenarios.

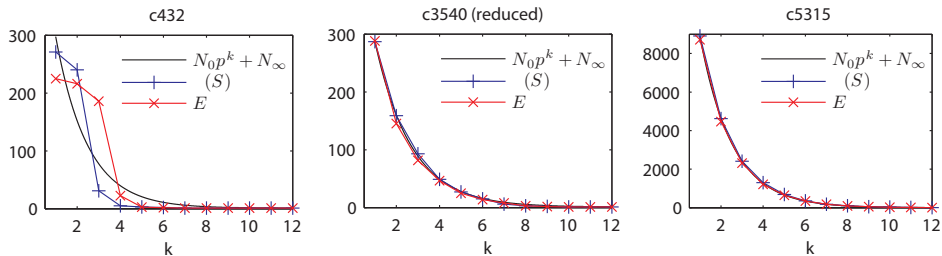

Figure 9: Actual number of remaining MC diagnoses $\Omega(S)$, expected number of remaining MC diagnoses $E$, and a geometric decay fit to $\Omega(S)$, stationary inputs, FRACTAL$^P$

Each plot in Fig. 9 shows a single probing session with a single initial observation. Figure 10 shows the goodness-of-fit criterion $R^2$ vs. the decay rate constant $p$ for all 100 observations and each of the 10 multiple runs of the Fig. 9 circuits.

It is visible from Fig. 10 that the absolute values of $R^2$ are (in most of the cases) close to 1. This is an indicator that the probing experiments fit the geometric decay model given in Eqn. 10 well. Figure 10 shows a "bad" topology (c432 on the left), and a "good" topology (c5315 on the right) that achieves decay rate $p$ close to 0.5 ($0.38 < p < 0.58$) with very high accuracy of the fit ($0.9896 \leq R^2 \leq 1$).

The expected number of remaining MC diagnoses is a good predictor of the actual number of MC diagnoses for all ISCAS85 circuits, as is shown in Table 11. The absolute values, again

---

5. There exist multi-probe generalizations of probing (de Kleer et al., 1992).

328

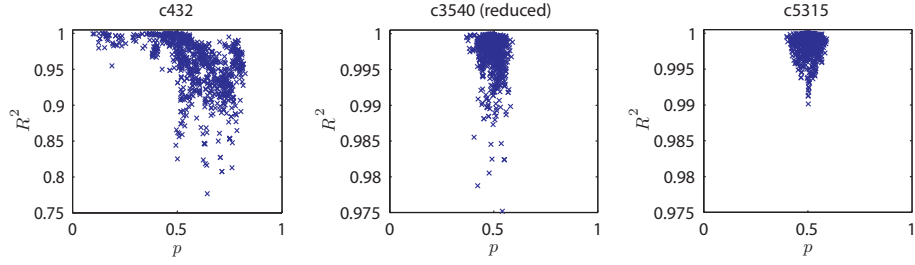

Figure 10: Geometric decay rate vs. goodness-of-fit for $\Omega(S)$, FRACTAL$^P$

depend on the topology, and we can see a smaller correlation $\rho$ for some c432 observations. In most of the cases, however, the correlation is significant, e.g., for all circuits and observations except c432 we have $\rho > 0.95$.

Table 11: Linear correlation coefficient $\rho$ of the expected number of remaining MC diagnoses $E$ and the actual number of remaining diagnoses $\Omega^{\leq}(S)$, stationary inputs, FRACTAL$^P$

| Name | Original | | | | Reduced | | | |
|---|---|---|---|---|---|---|---|---|
| | $\rho_{\min}$ | $\rho_{\text{avg}}$ | $\rho > 0.95$ | $\rho > 0.975$ | $\rho_{\min}$ | $\rho_{\text{avg}}$ | $\rho > 0.95$ | $\rho > 0.975$ |
| 74182 | 0.83 | 0.95 | 0.64 | 0.6 | 1 | 1 | 1 | 1 |
| 74L85 | 0.77 | 0.97 | 0.87 | 0.67 | 0.83 | 0.99 | 0.92 | 0.87 |
| 74283 | 0.97 | 0.99 | 1 | 1 | 0.83 | 0.98 | 0.83 | 0.76 |
| 74181 | 0.96 | 0.99 | 1 | 0.97 | 0.92 | 0.99 | 0.95 | 0.92 |
| c432 | 0.66 | 0.97 | 0.83 | 0.67 | 0.62 | 0.96 | 0.76 | 0.62 |
| c499 | 0.97 | 1 | 1 | 1 | 0.91 | 0.98 | 0.87 | 0.76 |
| c880 | 0.98 | 1 | 1 | 1 | 0.92 | 0.99 | 0.99 | 0.96 |
| c1355 | 0.99 | 1 | 1 | 1 | 0.86 | 0.98 | 0.88 | 0.79 |
| c1908 | 0.98 | 1 | 1 | 1 | 0.65 | 0.97 | 0.86 | 0.68 |
| c2670 | 0.98 | 1 | 1 | 1 | 0.7 | 0.96 | 0.72 | 0.55 |
| c3540 | 0.97 | 1 | 1 | 1 | 0.97 | 1 | 1 | 1 |
| c5315 | 0.99 | 1 | 1 | 1 | 0.7 | 0.98 | 0.91 | 0.81 |
| c6288 | 0.92 | 1 | 1 | 1 | 0.98 | 1 | 1 | 1 |
| c7552 | 0.95 | 1 | 1 | 0.99 | 0.82 | 0.96 | 0.7 | 0.51 |

In the second and third columns of Table 11 we can see the minimum and average correlations between $E$ and $\Omega^{\leq}(S)$. The third and fourth cases specify the fraction of observations for which we have $\rho > 0.95$ and $\rho > 0.975$, respectively. Columns 6 – 9 repeat this data for the reduced 74XXX/ISCAS85 circuits.

Table 12 summarizes the decay rate $p$ and the goodness-of-fit criterion $R^2$ for all observations and circuits. For c432, the values of $p$ and $R^2$ are more dispersed, while in the other experiments $p$ strongly resembles that of "chained-elements" (i.e., $p$ is close to 0.5). The minimum, maximum and average values of $p$ (per circuit) are given in columns $p_{\min}$, $p_{\max}$, and $p_{\text{avg}}$, respectively.

Table 12: Decay rate $p$ (minimal, maximal, and average) and goodness-of-fit $R^2$ (average) of geometric decay best-fit to $\Omega(S)$, stationary inputs, FRACTAL$^P$

|  | Original | | | | Reduced | | | |
| Name | $p_{\min}$ | $p_{\max}$ | $p_{\text{avg}}$ | $R^2_{\text{avg}}$ | $p_{\min}$ | $p_{\max}$ | $p_{\text{avg}}$ | $R^2_{\text{avg}}$ |
| --- | --- | --- | --- | --- | --- | --- | --- | --- |
| 74182 | 0.26 | 0.64 | 0.54 | 0.95 | 0.5 | 0.5 | 0.5 | 1 |
| 74L85 | 0.21 | 0.7 | 0.52 | 0.97 | 0.25 | 0.55 | 0.45 | 0.97 |
| 74283 | 0.31 | 0.64 | 0.49 | 0.99 | 0.4 | 0.58 | 0.49 | 0.96 |
| 74181 | 0.3 | 0.66 | 0.5 | 0.99 | 0.27 | 0.56 | 0.42 | 0.99 |
| c432 | 0.1 | 0.82 | 0.58 | 0.96 | 0.11 | 0.84 | 0.55 | 0.95 |
| c499 | 0.4 | 0.57 | 0.5 | 1 | 0.25 | 0.6 | 0.46 | 0.98 |
| c880 | 0.36 | 0.61 | 0.51 | 1 | 0.2 | 0.67 | 0.46 | 0.99 |
| c1355 | 0.39 | 0.6 | 0.51 | 1 | 0.25 | 0.59 | 0.46 | 0.98 |
| c1908 | 0.39 | 0.58 | 0.5 | 1 | 0.13 | 0.81 | 0.55 | 0.96 |
| c2670 | 0.37 | 0.6 | 0.51 | 1 | 0.22 | 0.85 | 0.65 | 0.89 |
| c3540 | 0.38 | 0.58 | 0.5 | 1 | 0.37 | 0.59 | 0.49 | 1 |
| c5315 | 0.4 | 0.59 | 0.5 | 1 | 0.18 | 0.89 | 0.52 | 0.96 |
| c6288 | 0.92 | 1 | 1 | 1 | 0.98 | 1 | 1 | 1 |
| c7552 | 0.95 | 1 | 1 | 0.99 | 0.82 | 0.96 | 0.7 | 0.51 |
| Average | 0.41 | 0.69 | 0.58 | 0.99 | 0.35 | 0.71 | 0.55 | 0.94 |

### 6.4 Experimental Summary

If we compare Table 6 and Table 11 we can see that the average correlation $\rho_{\text{avg}}$ decreases significantly. Hence, assuming limited observability (i.e., assuming that not all internals are measurable) decreases the quality of $E$ as a predictor of $\Omega(S)$. The increased statistical dispersion of $\rho$ is visible from the increased range $\rho_{\max} - \rho_{\min}$ (cf. Table 6, where $\rho_{\max}$ is always 1). For example, if we consider c2670, the standard deviation of all $E$ vs. $\Omega(S)$ correlation coefficients $\rho$ is $\sigma_\rho = 0.0031$ for FRACTAL$^P$ and $\sigma_\rho = 0.0783$ for FRACTAL$^{\text{ATPG}}$. The difference in dispersion of correlation coefficients is significant for all circuits, with smallest values for c432, where it is 0.0038 for FRACTAL$^P$ and 0.0825 for FRACTAL$^{\text{ATPG}}$.

By comparing Table 7, Table 9, and Table 12 we can see that the mean decay rates of FRACTAL$^{\text{ATPG}}$, FRACTAL$^G$, and FRACTAL$^P$ are similar (the average $p$ of FRACTAL$^G$ is 0.7 while the average $p$ of FRACTAL$^{\text{ATPG}}$ is 0.73). The average goodness-of-fit criterion $R^2$ for exponential decays is always good (0.88 for FRACTAL$^G$, 0.84 for FRACTAL$^{\text{ATPG}}$), and almost perfect in probing (0.97).

The summary of our experiments is best shown in Fig. 11. To factor out sampling error and to be able to perform exhaustive computations, we have chosen the smallest 74182 circuit. The original 74182 (a 4-bit carry-lookahead generator) has 19 components, 9 inputs, and 5 outputs. We have turned four of the inputs into controls (hence, $|\text{IN}| = 4$ and $|\text{CTL}| = 4$).

We have considered a random control policy in addition to FRACTAL$^P$, FRACTAL$^{\text{ATPG}}$, and FRACTAL$^G$. With a random control policy, at each step, a random value is assigned to each control variable. We have also shown an exhaustive control search where the expected number of remaining MC diagnoses is computed at each step, and for each possible control combination. This works with 74182 but leads to a combinatorial blow-up with any other (larger) circuit.

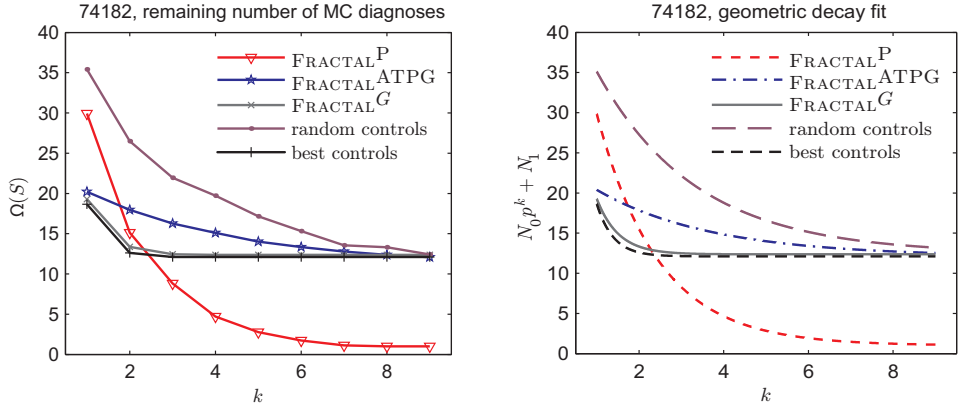

Figure 11: Comparison of all control policies

To reduce the stochastic error when plotting Fig. 11, we have replaced the sampling (for computing an expected number of remaining MC diagnoses) with an exhaustive method; this is possible as $|\text{IN}| = 5$. The only randomized decision is to choose the actual fault from the initial ambiguity group. To reduce the error due to this stochastic fault injection, we have tested each of the 5 control policies 100 times.

We can see in Fig. 11 that the least informed control policy (the random control policy simply does not use $E$) shows the worst decay in the number of remaining diagnoses. On the other extreme, the exhaustive control policy achieves the best decay. The price for this policy in terms of computational effort, however, is prohibitive. FRACTAL$^G$ achieves decay rates comparable to the exhaustive policy with affordable average-case complexity. FRACTAL$^{ATPG}$ has better complexity than FRACTAL$^G$, but the whole decay rate curve of FRACTAL$^{ATPG}$ is bounded from below by the one computed by FRACTAL$^G$.

Probing does not compare to active testing as both approaches have different assumptions on the observability of the model. Figure 11 shows the decay rate of probing to illustrate the different decay curves depending on the observability assumptions. In this experiment the probing decay rate geometric fit with $p = \frac{1}{2}$ almost perfectly fits the actual number of remaining MC diagnoses.

## 7. Conclusions

We have devised an algorithm, FRACTAL$^G$, for active testing that is (1) computationally efficient and (2) rapidly reduces the diagnostic uncertainty (measured as the number of remaining MC diagnoses) by manipulating a set of control variables. As fully optimizing (2) leads to a combinatorial blow-up, FRACTAL$^G$ achieves a compromise between (1) and (2) by using a greedy approximation approach for searching over the space of control assignments and a stochastic sampling method for computing the number of remaining MC diagnoses. The result is a fast algorithm (optimizing a whole FRACTAL scenario takes between 1 s

331

for 74182 and 40 min for c6288) that decreases the diagnostic uncertainty according to a geometric decay curve. This geometric decay curve fits the Fractal data well (the goodness-of-fit criterion $R^2$ is 0.88 on average) and provides steep decay (the average decay rate $p$ is 0.7).

We have applied Fractal$^G$ to the real-world problem of reducing the diagnostic uncertainty of a heavy-duty printer (Feldman, 2010). For that purpose, we have modeled the Paper Input Module (PIM). In the PIM case-study, Fractal$^G$ computed the most informative tests in troubleshooting multiple sensor and component failures. This happens even with a coarse-grained device model (only a few constraints per component), which shows an unexpected benefit of Fractal: trade-off of modeling complexity vs. test effort.

The optimality of Fractal$^G$ depends on the topology of and constraints on the input model. We can create models leading to arbitrarily bad optimality of Fractal$^G$ by, for example, directly encoding truth tables in SD. In practical situations, however, controls are independent. That means that applying a single control rarely "undoes" the effect of the previous ones. This also happens when arbitrary inputs are converted to controls, as in our experimentation benchmark. Consider, for example, a multiplier (c6288). Leaving out some of the inputs leads to "don't cares" in the output and hence some components (full-adders, and-gates) will remain untested. Subsequently assigning values to these left-out inputs will unambiguously exonerate or blame these untested components, which will help narrowing down the set of diagnostic hypotheses.

The most important benefit in applying Fractal to industrial cases is that active testing "trade-offs" modeling fidelity for computational complexity and extra testing. This enables users to achieve good diagnostic certainty without the large cost traditionally associated with developing high fidelity models based on physics of failure and other precision approaches.

We have compared the optimality and performance of Fractal$^G$ to an ATPG-based algorithm for sequential diagnosis, Fractal$^{ATPG}$. While the average decay rate of both algorithms is similar (average $p$ of Fractal$^{ATPG}$ is 0.73), the average goodness-of-fit criterion $R^2$ of Fractal$^{ATPG}$ is lower (0.84), which means that Fractal$^G$ is consistently closer to the optimal solution than is Fractal$^{ATPG}$. Fractal$^G$ has achieved better exponential decay compared to all algorithms except exhaustive control search. For example, the difference in the decay rate $p$ between Fractal$^G$ and exhaustive search for 74182 is 5.4%. The exhaustive control approach, however, takes minutes to complete even for a circuit as simple as 74182, and times-out with any model having more than 20 controls. As a result, we can conclude that Fractal$^G$ trades off a small decrease in $p$ for a significant performance speedup.